\documentclass[review,12pt,3p]{elsarticle}

\usepackage{titlesec}

\titleformat{\paragraph}
{\normalfont\normalsize}{\theparagraph}{1em}{}
\titlespacing*{\paragraph}
{0pt}{3.25ex plus 1ex minus .2ex}{1.5ex plus .2ex}

\usepackage{times}
\usepackage{balance}
\usepackage{setspace}
\usepackage{algorithm}
\usepackage{algorithmic}
 \usepackage{relsize}

\usepackage{boxedminipage} 
\usepackage{graphicx}
\usepackage{verbatim} 
\usepackage{amsmath}
\usepackage{times} 
\usepackage{gensymb}
\usepackage{dsfont}
\usepackage{multirow}
\usepackage{float}
\usepackage{amsfonts}
\usepackage{epstopdf}
\usepackage{mathrsfs}
\usepackage[T1]{fontenc}
\usepackage{url}
\usepackage{amssymb}
\usepackage[tight,footnotesize]{subfigure}
\usepackage[caption=false,font=footnotesize]{subfig}
\usepackage{epsfig}

\begin{document}

\begin{abstract}		

Dynamic ensemble selection (DES) techniques work by estimating the competence level of each classifier from a pool of classifiers, and selecting only the most competent ones for the classification of a specific test sample. The key issue in DES is defining a suitable criterion for calculating the classifiers' competence. There are several criteria available to measure the level of competence of base classifiers, such as local accuracy estimates and ranking. However, using only one criterion may lead to a poor estimation of the classifier's competence. In order to deal with this issue, we have proposed a novel dynamic ensemble selection framework using meta-learning, called META-DES. A meta-classifier is trained, based on the meta-features extracted from the training data, to estimate the level of competence of a classifier for the classification of a given query sample. An important aspect of the META-DES framework is that multiple criteria can be embedded in the system encoded as different sets of meta-features. However, some DES criteria are not suitable for every classification problem. For instance, local accuracy estimates may produce poor results when there is a high degree of overlap between the classes. Moreover, a higher classification accuracy can be obtained if the performance of the meta-classifier is optimized for the corresponding data. In this paper, we propose a novel version of the META-DES framework based on the formal definition of the Oracle, called META-DES.Oracle. The Oracle is an abstract method that represents an ideal classifier selection scheme. A meta-feature selection scheme using an overfitting cautious Binary Particle Swarm Optimization (BPSO) is proposed for improving the performance of the meta-classifier. The difference between the outputs obtained by the meta-classifier and those presented by the Oracle is minimized. Thus, the meta-classifier is expected to obtain results that are similar to the Oracle. Experiments carried out using 30 classification problems demonstrate that the optimization procedure based on the Oracle definition leads to a significant improvement in classification accuracy when compared to previous versions of the META-DES framework and other state-of-the-art DES techniques.

\end{abstract} 


\begin{frontmatter} 

\title{META-DES.Oracle: Meta-learning and feature selection for dynamic ensemble selection}

\author[ets]{Rafael M. O. Cruz\corref{corr1}}
\ead{rmoc@cin.ufpe.br}

\author[ets]{Robert Sabourin}
\ead{robert.sabourin@etsmtl.ca}

\author[ufpe]{George D. C. Cavalcanti}
\ead{gdcc@cin.ufpe.br}

\address[ets]{LIVIA, \'{E}cole de Technologie Sup\'{e}rieure, University of Quebec, Montreal, Que., Canada - www.livia.etsmtl.ca
			 }
\address[ufpe]{Centro de Inform\'{a}tica, Universidade Federal de Pernambuco, Recife, PE, Brazil - www.cin.ufpe.br/$\sim$viisar\\
       }
			
\cortext[corr1]{Corresponding author. Email Address: rafaelmenelau@gmail.com.}

\end{frontmatter}

\section{Introduction}

Multiple Classifier Systems (MCS) aim to combine classifiers in order to increase the recognition accuracy in pattern recognition systems~\cite{kittler,kuncheva,wozniak2013hybrid,wozniak2014survey}. MCS are composed of three phases~\cite{Alceu2014}: (1) Generation, (2) Selection, and (3) Integration. In the first phase, a pool of classifiers is generated. In the second, a single classifier or a subset having the best classifiers of the pool is(are) selected. We refer to the subset of classifiers as the Ensemble of Classifiers (EoC). In the last phase, called integration, the predictions of the selected classifiers are combined to obtain the final decision.

The classifier selection phase can be either static or dynamic. In static selection, the ensemble is selected during the training stage. The classifiers with the best performance, according to the selection criteria, considering the whole training or validation distribution are selected to compose the ensemble. Then, the ensemble is used for the classification of all unseen data. In dynamic approaches, the ensemble of classifiers is selected during the test phase. For each test sample, the competence of the base classifiers is estimated according to a selection criterion. Then, only the classifier(s) that attain a certain competence level, are used to predict the label of the given test sample. Recent works in the MCS literature have shown that dynamic ensemble selection (DES) techniques achieve higher classification accuracy when compared to static ones~\cite{Alceu2014,CruzPR,knora}. This is especially true for ill-defined problems, i.e., for problems where the size of the training data is small, and there are not enough data available to train the classifiers~\cite{paulo2,logid}. Moreover, using dynamic ensemble selection, we can solve classification problems with a complex non-linear decision boundary using only a few linear classifiers, while static ensemble techniques, such as Bagging and AdaBoost, cannot~\cite{reportarXiv}. 

When dealing with DES, the key issue is to define a suitable criterion to select the most competent classifiers to predict the label of a specific query sample. Several criteria have previously been proposed, based on different sources of information, such as the classifier local accuracy estimates in small regions of the feature space surrounding the query instance, called the region of competence~\cite{lca,knora}, probabilistic models~\cite{Woloszynski,WoloszynskiKPS12,Kurzynski2010}, ranking~\cite{classrank} and classifier behavior~\cite{mcb,paulo2}. In our previous work~\cite{CruzPR}, we proposed a novel DES framework using meta-learning, called META-DES. The framework is divided into three steps: (1) Overproduction, where the pool of classifiers is generated; (2) Meta-training, where the meta-features are extracted using the training data, and used as inputs to train a meta-classifier that works as a classifier selector; and (3) the Generalization phase, in which the meta-features are extracted from each query sample and used as input to the meta-classifier. The meta-classifier decides whether the base classifier is competent enough to classify the test sample. 

The main advantage of the META-DES framework is its modularity. Any criterion used to estimate the level of competence of base classifiers can be encoded as a new set of meta-features and added to the system. A total of five sets of meta-features were proposed in~\cite{CruzPR}, each one representing a different DES criterion, such as local accuracy information and degree of confidence. Moreover, in~\cite{reportarXiv}, a case study is presented demonstrating how the use of multiple criteria leads to a more robust dynamic selection technique. Using multiple sets of meta-features, even though one criterion might fail due to imprecisions in the local regions of the feature space or due to low confidence results, the system can still achieve a good performance as other meta-features are considered by the selection scheme. Since the META-DES framework considers the dynamic selection problem as a meta-classification problem, we can significantly improve the recognition accuracy of the system by focusing only on optimizing the performance of the meta-classifier.

However, there are some drawbacks to the META-DES framework. First, there are different sources of information that were not considered by the previous version of the system, such as probabilistic models, ambiguity, and ranking. Secondly, all sets of meta-features are used for every classification problem with no pre-processing step at all. As stated by the ``No Free Lunch'' theorem~\cite{freelunch}, there is no criterion for dynamic selection that outperforms all others over all possible classes of problems. Different classification problems may require distinct sets of meta-features. The meta-classifier training process is not optimized for each classification problem. This can also lead to low classification results, since we found that the training of the meta-classifier is problem-dependent~\cite{icpr2014}. For these reasons, the results obtained by the META-DES framework were still far from those achieved by the Oracle. The Oracle is an abstract model defined in~\cite{Kuncheva:2002}, which always selects the classifier that predicted the correct label, for the given query sample, if such a classifier exists. Although it is possible to achieve results higher than the Oracle by working on the supports given by the base classifier~\cite{wozniak2010designing,wozniak2014survey}, from a dynamic selection point of view, the Oracle represents the perfect dynamic selection scheme, since it always selects the classifiers that predict the correct label~\cite{DidaciGRM05}. As stated by Ko et al.~\cite{knora}, to achieve better results using dynamic selection methods, we need to better understand the behavior of the Oracle. However, addressing its behavior is more complex than applying a single selection criteria, since distinct classification problems may require the use of different selection criteria as they associated with distinct degrees of data complexity~\cite{HoB02}.


In this paper, we propose a new optimization scheme to the META-DES framework in order to better address the behavior of the Oracle. In the first stage, a pool of linear classifiers is generated using the Bagging technique~\cite{bagging}. In this case, the Perceptron classifier is considered as the base classifier model, since we demonstrated in~\cite{reportarXiv} that using dynamic selection it is possible to solve non-linear classification problems with complex decision boundaries, using a pool containing only five linear base classifiers. Even though the individual accuracy of each base classifier is approximately 50\%, the selection mechanism embedded in the framework is able to select the most competent ones for the classification of a given query instance. 

In the second stage, 15 sets of meta-features are proposed, using sources of information that were not explored in the previous version framework, such as ranking, ambiguity and probabilistic models applied over the supports obtained by the meta-classifier, for a better estimation of the competence level of the base classifiers. The additional meta-features are motivated by a recent analysis conducted in~\cite{Cruz2014ANNPR}, demonstrating that using different sources of information to estimate the competence level of the base classifiers leads to a more robust DES technique.
The meta-features are used as input to a meta-classifier that is trained to identify whether or not a base classifier is competent enough for the classification of an input sample.

Following that, a meta-feature selection scheme is applied in order to optimize the performance of the meta-classifier, based on a formal definition of the Oracle. A Binary Particle Swarm Optimization (BPSO) using a V-shaped and S-shaped transfer function~\cite{MirjaliliL13} is used in the optimization process. The difference between the level of competence estimated by the meta-classifier and that estimated by the Oracle is used as the fitness function for the BPSO. In other words, the optimization scheme seeks a meta-features vector that minimizes the difference between the behavior of the meta-classifier and that of the Oracle in estimating the competence level of the base classifiers. Thus, the meta-classifier is more likely to present results that are closer to that of the Oracle. We call the proposed system META-DES.Oracle, since the formal definition of the Oracle is used during the training stage of the meta-classifier. 

The classification stage is performed using a hybrid dynamic selection and weighting scheme. First, the meta-classifier is used to estimate the competence level of each base classifier. The classifiers that attain a certain level of competence are selected to compose the ensemble. Next, the meta-classifier is used to compute the weights of the selected base classifiers to be used in a weighted majority voting scheme. In this way, the base classifiers that present a higher level of competence have greater influence on the ensemble decision. 

Experiments are conducted over 30 classification problems derived from different data repositories. We compare the results obtained by the proposed META-DES.Oracle with 10 state-of-the-art dynamic selection techniques,as well as static ensemble methods (e.g., AdaBoost~\cite{boosting} and Random Forests~\cite{breiman2001random,rokach2016decision}) and single classifier models, such as Support Vector Machines (SVM) with Gaussian Kernel, Multi-Layer Perceptron (MLP) Neural Network. The goal of the experimental study is to answer the following research questions: (1) Are different sets of meta-features better suited for different problems? (2) Are all 15 sets of meta-feature relevant? (3) Does the META-DES.Oracle obtain a significant gain in classification accuracy when compared to the previous versions of the META-DES framework? (4) Does the META-DES.Oracle outperform state-of-the-art DES techniques? (5) Is the performance obtained by the proposed framework comparable to that of the best families of classifiers in the literature~\cite{delgado14a}?

In a nutshell, the contributions of this work are: (1) A novel DES framework based on meta-learning which selects the best set of meta-features in order mimic the selection mechanism of the Oracle. (2) The definition of 15 sets of meta-features as well as categorization of several DS criteria based on their source of information. (3) A formal definition of the Oracle as the ideal classifier selection scheme. (4) Optimization of the META-DES framework based on the formal definition of the Oracle. (5) A extensive comparison among the proposed META-DES.Oracle with 10 state-of-the-art techniques as well as static ensemble and the best single classifier models based on~\cite{delgado14a}. As far as we know, this is the first paper in the dynamic selection literature that perform a comparison among several DS techniques and different classification schemes.

This paper is organized as follows: Section~\ref{sec:relatedWork} introduces state-of-the-art techniques for dynamic classifier and ensemble selection. The META-DES.Oracle is detailed in Section~\ref{sec:proposed}. In Section~\ref{sec:classifierCompetence}, we describe the 15 sets of meta-features proposed in this work. An illustrative example using synthetic data is shown in Section~\ref{sec:synthetic}. The experimental study is conducted in Section~\ref{sec:experiments}. Finally, our conclusion and future works proposals are given in the last section.

\section{Related Works}
\label{sec:relatedWork}

\subsection{Dynamic selection}
In static ensemble methods, such as in Decision Forests~\cite{rokach2016decision} and in Boosting methods~\cite{boosting}, the ensemble of classifiers is defined in the training phase, and is used to predict the label of all test samples during the generalization phase. In contrast, dynamic ensemble selection techniques~\cite{classrank,knora,docs,paulo2,Woloszynski,lca,vriesmann2015combining,cruz2016prototype} consists of, based on a pool of classifiers $C$, finding a single classifier $c_{i}$ or an ensemble of classifiers $C'$ that has the most competent classifiers to predict the label for a specific test sample, $\mathbf{x}_{j}$. 
The ensemble is selected in a dynamic fashion according to each new test sample. This property makes dynamic ensemble selection techniques a robust approach to deal with many pattern recognition applications, such as, handwritten recognition~\cite{knora}, face recognition~\cite{bashbaghi2016robust}, remote sensing image classification~\cite{Smits_2002}, offline signature verification~\cite{batista2012dynamic} and the recognition of EMG signals in a bio-prosthetic hand~\cite{kurzynski2011dynamic}. 
 
In addition, recent works have demonstrated that dynamic selection techniques can also be used in different classification contexts. For instance, in one-class classification~\cite{Krawczyk201643}, where the system has no access to counterexamples during the training stage, and may require to select the most appropriate classifiers on-the-fly. Another context were dynamic selection has shown some success is in One-Versus-One (OVO) decomposition strategies~\cite{Galar20111761}. OVO works by dividing a multi-class classification problem into as many binary problems as all possible combinations between pair of classes~\cite{Galar20111761}. Each base classifier is trained solely to distinguish between each pair of classes. When a new query sample is presented for classification, the outputs of all base classifiers are combined to predict its label. The problem of OVO strategies relies on the fact that each base classifier is only trained to distinguish between two classes. Not all base classifiers are competent to classify the query sample, since they might not even be trained for the corresponding pair of classes. The outputs of such non-competent classifiers may hinder the performance of the system~\cite{Galar20133412}. 

Galar et al.~\cite{Galar20133412} proposed the Dynamic-OVO strategy, which applies a dynamic selection mechanism in order to avoid non-competent classifiers to weight in the ensemble decision. In this strategy, the neighborhood of the query instance is computed using the K-Nearest Neighbors method. Only the classifiers that were trained considering the classes present in the neighborhood of the query sample are used in the combination scheme. An updated version of the Dynamic-OVO, the Distance-based Relative Competence Weighting combination (DRCW-OVO) was proposed in~\cite{Galar201528} to further reduce the impact of non-competent classifiers using a weighting mechanism. The outputs of the selected classifiers are weighted depending on the closeness of the query instance to the nearest neighbors of each class in the problem. The larger the distance is, the lower weight the classifier, has and vice versa~\cite{Galar201528}. Another interesting strategy is the DYNOVO technique~\cite{Mendialdua2015298}. This method performs dynamic classifier selection in each sub-problem of the OVO decomposition, and select the best base classifiers to classify the query sample. In this case, an adaptation of the Overall Local Accuracy (LCA)~\cite{lca} strategy for OVO is proposed to estimate the competence of the base classifiers.

Nevertheless, the most important component of DES techniques is the criterion used to measure the level of competence of a base classifier $c_{i}$ for the classification of a given query sample $\mathbf{x}_{j}$. The most common approach involves estimating the accuracy of the base classifiers in small regions of the feature space surrounding the query sample, $\mathbf{x}_{j}$, called the region of competence. This region is usually defined based on the nearest neighbor rule applied to either the training~\cite{lca} or validation data~\cite{knora}. Based on the region of competence, there are several sources of information that can be used to measure the competence of the classifier in the DES literature~\cite{Alceu2014}: Measures based solely on accuracy, such as the Overall Local Accuracy (OLA)~\cite{lca}, Local Classifier Accuracy (LCA)~\cite{lca} and Modified Local Accuracy (MLA)~\cite{lca}, ranking information such as the Classifier Rank~\cite{classrank} and the simplified classifier rank~\cite{lca}, probabilistic information calculated over the decision obtained by the base classifiers such as the Kullback Leibler divergence, DES-KL~\cite{WoloszynskiKPS12} and the randomized reference classifier DES-PRC~\cite{Woloszynski}, classifier behavior calculated using output profiles such as the KNOP technique~\cite{paulo2} and the KNORA family of techniques~\cite{knora} using Oracle information. Brun et al.~\cite{Brun2016} also presented the use of data complexity measures such as the Fisher's Discriminant Ratio~\cite{HoB02} to aid in the search for the most competent classifiers. Furthermore, there are some selection criteria that estimate the competence level of a whole ensemble of classifiers rather than the competence of each base classifier individually, such as the degree of consensus used in the Dynamic Overproduction and Choose technique (DOCS)~\cite{docs}, diversity~\cite{YasarSaglam2016,anne} and data handling~\cite{dceid}. 

An important concept in the DES literature is the definition of the Oracle. The Oracle is an abstract model defined in~\cite{Kuncheva:2002}, which always selects the classifier that predicted the correct label, for the given query sample, if such a classifier exists. In other words, it represents the ideal classifier selection scheme. The Oracle is used in the DES literature in order to determine whether the results obtained by the proposed DES techniques is close to ideal accuracy or whether there is still room for improvements. As reported in a recent survey~\cite{Alceu2014}, the results obtained by DES techniques based solely on one source of information are still far from those achieved by the Oracle. As stated by Ko et al.~\cite{knora}, addressing the behavior of the Oracle is much more complex than applying a simple neighborhood approach, and the task of figuring out its behavior based merely on the pattern feature space is not an easy one. In addition, in our previous work~\cite{ijcnn2011}, we demonstrated that the use of local accuracy estimates alone is insufficient to achieve good generalization performance.

To address these issues, in~\cite{CruzPR} we proposed a novel DES framework using meta-learning, called META-DES. From a meta-learning perspective, the dynamic selection problem can be seen as another classification problem, called the meta-problem. This meta-problem uses different criteria regarding the behavior of a base classifier in order to decide whether or not a base classifier $c_{i}$ is competent enough to classify a given sample $\mathbf{x}_{j}$. In this paper, our aim therefore is to optimize the performance of the meta-classifier, using the meta-classification environment, to obtain results closer to those of the Oracle. 

\subsection{Feature selection using Binary Particle Swarm Optimization (BPSO)}

Given a set of features $m$, the objective of feature selection is to identify the most informative subset of features $m^{'} \in m$. The reasons for using feature selection methods~\cite{KhushabaAA11} are: removal of redundant and irrelevant features, reduction of dimensionality, reduction of the computational complexity of the system, as well as improvement of the classification accuracy. There are two main factors when dealing with feature selection: the evaluation method, which is applied to compute the fitness of each solution, and the search strategy, which is used to explore the feature space in the search for a more suitable subset of features. 

For the search strategy, the recent focus in the feature selection literature has been on evolutionary computation techniques, such as Genetic Algorithms (GA)~\cite{haupt2004practical,RadtkeWS06}, Particle Swarm Optimization (PSO)~\cite{Kennedy:PSO,MirjaliliL13,ChuangTY11}, Differential Evolution (DE)~\cite{Price:2005,al2013feature,MirjaliliL13} and Ant Colony Optimization (ACO)~\cite{Dorigo:2004}. Evolutionary computation techniques have been shown to outperform other feature selection methods, such as sequential feature selection SFS, in many applications, especially when dealing with larger feature vectors, i.e., for classification problems with more than 50 features~\cite{KudoS00}.

Particle Swarm Optimization (PSO) is an evolutionary computation technique inspired from the social behavior of birds flocking~\cite{Kennedy:PSO}. PSO is one of the most used evolutionary algorithms, due to its simplicity and low computational cost. The technique is based on a group of particles flying around in the search space to find the best solution. Recent works have shown the preference for PSO over other classical optimization techniques, such as GA because GA has too many parameters to set. Moreover, GA is very sensitive to the probability of crossover and mutation operators, as well as to the initial population of solutions. Therefore, it is likely to get stuck into local minima~\cite{KhushabaAA11}. For this reason, BPSO has been shown to outperform other optimization algorithms in performing feature selection~\cite{Kennedy:PSO,Chuang:2008,FirpiG04}.

\section{The META-DES.Oracle}
\label{sec:proposed}

The META-DES framework is based on the assumption that the dynamic ensemble selection problem can be considered as a meta-problem~\cite{icpr2014}. This meta-problem uses different criteria regarding the behavior of a base classifier $c_{i}$, in order to decide whether it is competent enough to classify a given test sample $\mathbf{x}_{j}$. The meta-problem is defined as follows~\cite{CruzPR}:

 \begin{itemize}
 
 \item The \textbf{meta-classes} are either ``competent'' (1) or ``incompetent'' (0) to classify $\mathbf{x}_{j}$.
 
 \item Each set of \textbf{meta-features} $f_{i}$ corresponds to a different criterion for measuring the level of competence of a base classifier.
 
 \item The meta-features are encoded into a \textbf{meta-features vector} $v_{i,j}$.
 
 \item A \textbf{meta-classifier} $\lambda$ is trained based on the meta-features $v_{i,j}$ to predict whether or not $c_{i}$ will achieve the correct prediction for $\mathbf{x}_{j}$, i.e., if it is competent enough to classify $\mathbf{x}_{j}$.
 
 \end{itemize}

\begin{figure*}[!ht]
  
   \begin{center}  	 
       	  \includegraphics[clip=,  width=1.000\textwidth]{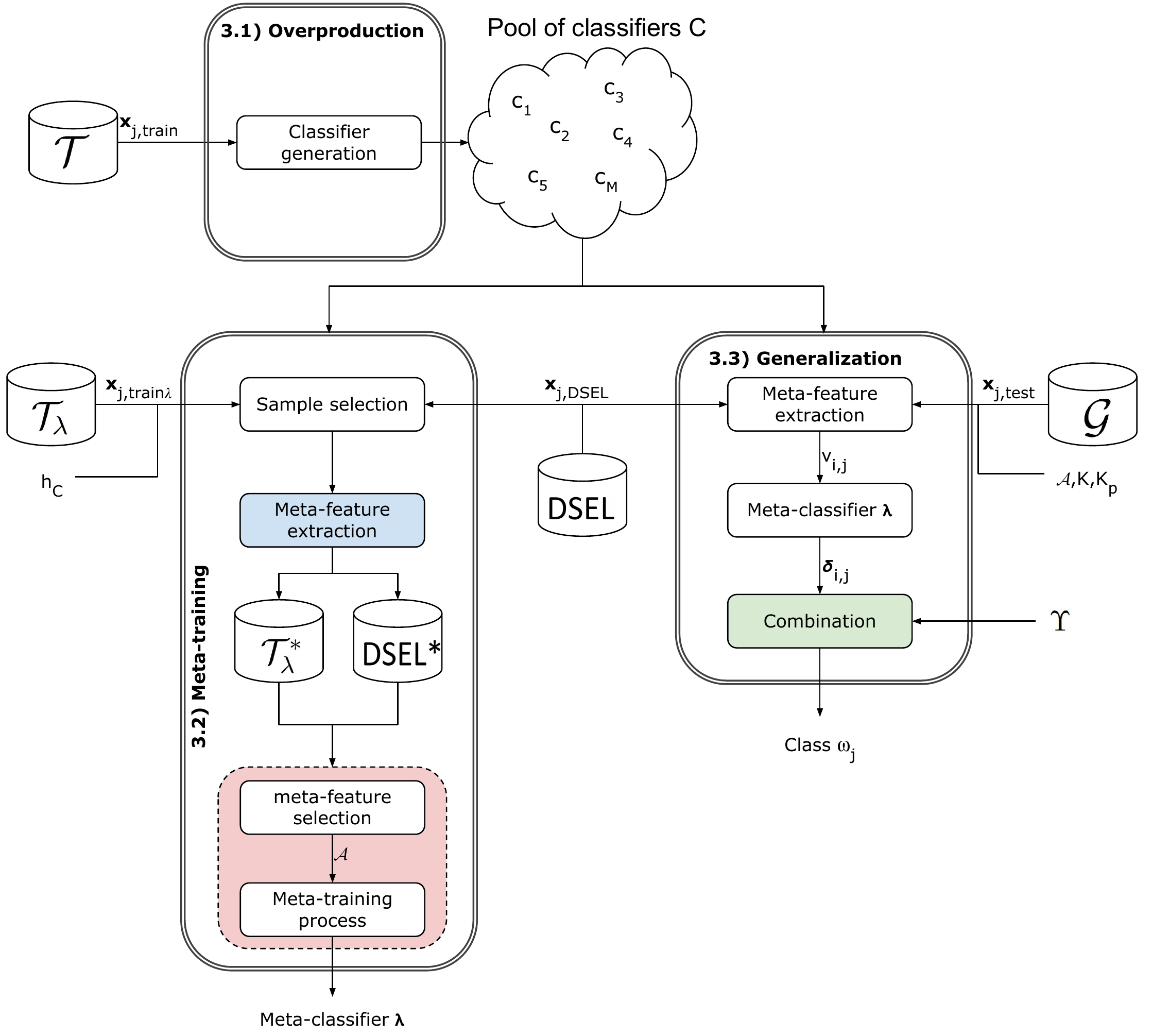}
   \end{center}
   
\caption{Overview of the proposed framework. It is divided into three steps: 1) Overproduction, where the pool of classifiers $C = \{c_{1}, \ldots, c_{M}\}$ is generated, 2) The training of the selector $\lambda$ (meta-classifier), and 3) The generalization phase, where the level of competence $\delta_{i,j}$ of each base classifier $c_{i}$ is calculated specifically for each new test sample $\mathbf{x}_{j,test}$. $h_{C}$, $K$, $K_{p}$ and $\Upsilon$ are the hyper-parameters required by the proposed system.}  

\label{fig:overview}
\end{figure*}

An overview of the META-DES framework is illustrated in Figure~\ref{fig:overview}. The framework is divided into three phases: (1) Overproduction, (2) Meta-training, and (3) Generalization. Phases (1) and (2) are performed in offline mode, i. e., during the training stage of the framework. In the overproduction phase, the pool of classifiers $C$ is generated using the training set $\mathcal{T}$. The following step is the meta-training stage, in which the meta-features are extracted for the training of the meta-classifier $\lambda$. In this stage, the meta-features are extracted from the meta-training set, $\mathcal{T}_{\lambda}$, and from the dynamic selection dataset, $DSEL$. The meta-data extracted from $\mathcal{T}_{\lambda}$, denoted by $\mathcal{T}_{\lambda}^{*}$, are used for the training of the meta-classifier, and those extracted from $DSEL$, denoted by $DSEL^{*}$, are used as validation data during the BPSO optimization process.  Phase (3) is conducted on-the-fly, with the arrival of each new test sample, $\mathbf{x}_{j,test}$, coming from the generalization dataset $\mathcal{G}$. For each base classifier $c_{i}$, a meta-features vector $v_{i,j}$ is extracted, corresponding to the behavior of the base classifier $c_{i}$ for the classification of $\mathbf{x}_{j,test}$. $v_{i,j}$ is passed down to the meta-classifier $\lambda$ that estimates if $c_{i}$ is competent enough to predict the label for $\mathbf{x}_{j,test}$. After all the classifiers in the pool $C$ are evaluated, the selected classifiers $C'$ are combined using a weighted majority voting approach to predict the label $w_{l}$ of $\mathbf{x}_{j,test}$. The main changes to the META-DES framework proposed in this paper are highlighted in different colors:

\begin{enumerate}
\item  The meta-feature extraction process, in which 15 sets of meta-features are extracted. Ten new sets of meta-features are proposed in this work in order to explore different sources of information for estimating the competence level of the base classifiers, such as probabilistic models, ambiguity, behavior and ranking. The meta-feature extraction process is presented in Section~\ref{sec:classifierCompetence}.

\item  The meta-features selection using Binary Particle Swarm Optimization and guided by Oracle information for achieving a behavior closer to the Oracle. The meta-features selection process is detailed in Section~\ref{sec:featureSelection}. 

\item The combination approach, where a hybrid dynamic selection and weighting approach is considered for the classification of the query sample $\mathbf{x}_{j,test}$ (Section~\ref{sec:generalizationPhase}). 

\end{enumerate}

\subsection{Overproduction}

Similarly to~\cite{CruzPR}, the Overproduction phase is performed using the Bagging technique~\cite{bagging}. The Bagging technique works by randomly selecting different bootstraps of the data for training each base classifier $c_{i}$. Each bootstrap uses of 50\% of the training data. The pool of classifiers $C$ is composed of 100 linear Perceptrons for the two-class problems and 100 multi-class linear Perceptrons for the multi-class problems. The use of linear classifiers is motivated by the finding in~\cite{reportarXiv,AlpaydinJ96,KunchevaR07} showing that the META-DES framework can solve complex non-linear classification problems with complex decision boundaries using only linear classifiers~\cite{reportarXiv}. 

\subsection{Meta-training Phase}

In this stage, the meta-features are extracted for the training of the meta-classifier $\lambda$. In this version of the framework we extract meta-data from two sets: the meta-training set $\mathcal{T}_{\lambda}$ and the dynamic selection (validation) $DSEL$. The meta-data extracted from the set $\mathcal{T}_{\lambda}$, denoted by $\mathcal{T}_{\lambda}^{*}$ are used for the training of the meta-classifier. The meta-data extracted from the set $DSEL$, denoted by $DSEL^{*}$ are used as validation data in the BPSO optimization scheme for preventing overfitting. 

\subsubsection{Sample Selection}

The first step in the meta-data generation process is the sample selection mechanism. The sample selection mechanism is employed in order to focus the training of the meta-classifier to deal with cases in which the extent of consensus of the pool is low, i.e., when there is a disagreement between the classifiers in the pool, for the correct label. For each instance $\mathbf{x}_{j}$\footnote{$\mathbf{x}_{j,DSEL}$ coming from the set DSEL or $\mathbf{x}_{j,train_{\lambda}}$ coming from the set $\mathcal{T}_{\lambda}$} coming from either the meta-training set, $\mathcal{T}_{\lambda}$, or the dynamic selection dataset DSEL, the consensus of the pool is computed by the percentage of base classifiers in the pool that predicts its correct label, denoted by $H \left ( \mathbf{x}_{j}, C \right )$. If the percentage falls below the consensus threshold, $h_{c}$, the sample, $\mathbf{x}_{j}$, is passed down to the meta-features extraction process. 

Next, for each base classifier, $c_{i} \in C$,  15 sets of meta-features are computed. Each set of meta-features is detailed in Section~\ref{sec:classifierCompetence}. The meta-feature vector $v_{i,j}$ containing the 15 sets of meta-features is obtained at the end of the process. The meta-feature vector $v_{i,j}$ represents the behavior of the base classifier $c_{i}$ for the classification of the query sample $\mathbf{x}_{j}$. If the base classifier $c_{i}$ predicts the correct label for $\mathbf{x}_{j}$, the class attribute of $v_{i,j}$, $\alpha_{i,j} = 1$ ($v_{i,j}$ belongs to the meta-class ``competent''), otherwise $\alpha_{i,j} = 0$ (belongs to the meta-class ``incompetent''). $v_{i,j}$ is stored in either $\mathcal{T}_{\lambda}^{*}$ or DSEL$^{*}$. 


\subsubsection{Meta-Feature Selection Using Binary Particle Swarm Optimization (BPSO)}
\label{sec:featureSelection}
Since we are dealing with feature selection, a binary version of the PSO algorithm, BPSO is considered. BPSO has been shown in many applications to outperform other optimization algorithms in performing feature selection~\cite{Kennedy:PSO,Chuang:2008,FirpiG04}. There are many versions of the BPSO algorithm, such as the Improved BPSO~\cite{Chuang:2008}, CatfishBPSO~\cite{ChuangTY11} and MBPSO~\cite{WangWFZ08}. Mirjalili et al.~\cite{MirjaliliL13} shows that the most important factor for achieving good convergence and avoiding local minima is the transfer function, that is responsible for mapping the continuous search space into a binary space. Generally speaking, there are two main types of transfer functions, S-shaped and V-shaped~\cite{MirjaliliL13}. The main difference between the two families derives from the observation that the S-Shaped functions force the particles to switch 0 or 1 values at each generation, while the V-Shaped transfer functions encourage particles to stay in their current position when their velocity values are low, and switch the values only when the velocity is high. For these reasons, V-Shaped transfer functions were shown to be better both in terms of robustness to local minima and convergence speed. In this work, we consider one S-Shaped transfer function and one V-Shaped function, which presented the best overall performance, considering 25 benchmark functions~\cite{MirjaliliL13}. 

Each particle (solution) is composed of a binary string $\mathcal{S}_{i} = \left \{ \mathcal{S}_{i,1}, ..., \: \mathcal{S}_{i,D} \right \}$ ($D$ is the number of meta-features), where every bit $\mathcal{S}_{i,d}$ represents a single meta-feature. The value ``1'' means the meta-feature is selected and ``0'' otherwise.

At each generation, the velocity of the i-th particle is computed using Equation~\ref{eq:velocity}:

\begin{equation}
\label{eq:velocity}
velocity_{i}^{g + 1} = wv_{i}^{g} + c_{1} \times rand \times (pBest_{i} - \mathcal{S}_{i}^{g}) + c_{2} \times rand \times (gBest - \mathcal{S}_{i}^{g})
\end{equation}

\noindent Each particle makes use of its private memory, $pBest_{i}$, which represents the best position the i-th particle visited as well as the knowledge of the swarm, $gBest$, which represent the global best position visited, considering the whole swarm. The constant $w$ corresponds to the inertia weight, $c_{1}$ and $c_{2}$ are the acceleration coefficients, and $rand$ is a randomly generated number between $0$ and $1$. The term $c_{1} \times rand \times (pBest_{i} - \mathcal{S}_{i}^{g})$ represents the private knowledge of the i-th particle, and the term $c_{2} \times rand \times (gBest - \mathcal{S}_{i}^{g})$ represents the collaboration of particles.

When dealing with binary search spaces, updating the position of a particle means switching between ``0'' and ``1'', i.e., whether or not the meta-feature is selected. The switching is conducted based on the velocity of the particle. The higher the velocity of a particle, the higher its probability of changing positions should be. However, the velocities are computed in the real space rather than in the binary space (as shown in Equation~\ref{eq:velocity}). The velocity of the particle needs to be converted into a probability value, representing the probability of changing the position of the particle from ``$0$'' to ``$1$'' and vice versa. This step is conducted using a transfer function, $\mathscr{T}$. A transfer function should work in a way that the higher the velocity value, the higher the probability of changing position will be, since particles with higher velocity values are probably far from the best solutions ($pBest_{i}$ and $gBest$). Similarly, a transfer function must present a lower probability of switching position for lower velocity values~\cite{MirjaliliL13}. The position of the i-th particle is updated according to Equation~\ref{eq:switching}.

\begin{equation}
\label{eq:switching}
\mathcal{S}_{i}^{g+1} = \left\{\begin{matrix}
(\mathcal{S}_{i}^{g+1})^{-1} & If \; \;  rand < \mathscr{T}(velocity_{i}^{d}(g+1)) \\ 
\mathcal{S}_{i}^{g+1} & If \; \; rand \ge \mathscr{T}(velocity_{i}^{d}(g+1))
\end{matrix}\right.
\end{equation}

Generally speaking, there are two main types of transfer functions, S-shaped and V-shaped~\cite{MirjaliliL13}. In this work we consider one S-shaped transfer function proposed in~\cite{Kennedy:PSO} and one V-shaped transfer function proposed in~\cite{MirjaliliL13}, in Equations~\ref{eq:Sshaped} and~\ref{eq:Vshaped}, respectively. These transfer functions were selected since they obtained the best results in several optimization benchmarks~\cite{MirjaliliL13}. 

\begin{equation}
\label{eq:Sshaped}
\mathscr{T}_{S}(x) = \frac{1}{1 + e^{-2x}}
\end{equation}

\begin{equation}
\label{eq:Vshaped}
\mathscr{T}_{V}(x) = \left | \frac{2}{\pi} arc \; tan \left ( \frac{\pi}{2} x \right ) \right |
\end{equation}

\paragraph{Fitness Function - distance to the oracle}
\label{sec:fitness}

The optimization of the meta-classifier is conducted based on the definition of the Oracle. From the dynamic selection point of view, the Oracle is seen as the ideal dynamic selection technique, which always selects the classifier that predicts the correct label, $\mathbf{x}_{j}$, and rejects otherwise. From the classifier competence point of view, the selection mechanism employed by the Oracle as the ideal dynamic classifier selection scheme. In this work, we formalize the Oracle as an ideal selection scheme using Equation~\ref{eq:oracle}.


\begin{equation}
\label{eq:oracle}
\begin{cases}
\delta_{i,j}  = 1, & \text{ if } c_{i} \; \:  \text{correctly classifies} \;\:   \mathbf{x}_{j} \\ 
\delta_{i,j}  = 0, & \text{ otherwise} 
\end{cases}
\end{equation}

The level of competence $\delta_{i,j}$ of a base classifier $c_{i}$ is equals to $1$ if it predicts the correct label for $\mathbf{x}_{j}$, and $0$ otherwise. In the META-DES framework, we want the meta-classifier, $\lambda$ to perform similarly to the Oracle, in a way that it should identify which base classifiers in the pool is competent to predict the label of an  unknown test instance $\mathbf{x}_{j}$ and should be selected to compose the ensemble. In order to achieve such behavior, we measure the difference between the estimation of competence achieved by the ideal selection scheme,represented by the Oracle, and the estimation of competence obtained by the meta-classifier $\lambda$ in the fitness function of the optimization scheme.

The fitness function is computed as follows: Given that $\delta_{i,j}^{\lambda}$ and $\delta_{i,j}^{Oracle}$ are the level of competence of the base classifier $c_{i}$ for the classification of the instance, $\mathbf{x}_{j}$, computed by the META-DES framework and the Oracle, respectively. The difference between both techniques $d_{\lambda,Oracle}$ is calculated by the mean squared difference between their competence estimates, $\delta_{i,j}^{\lambda}$ and $\delta_{i,j}^{Oracle}$ (Equation~\ref{eq:dissimilarityDS}). 

\begin{equation}
\label{eq:dissimilarityDS}
d_{\lambda,Oracle} = \frac{1}{NM} \sqrt{ \sum_{j = 1}^{N}\sum_{i = 1}^{M} \left ( \delta_{i,j}^{\lambda} - \delta_{i,j}^{Oracle} \right )^{2} }
\end{equation}

\noindent where $N$ and $M$ are the size of the dataset and pool of classifiers, respectively. 

Therefore, the BPSO optimization searches for a meta-classifier which minimizes the distance $d_{\lambda,Oracle}$. In other words, we search for a meta-classifier $\lambda$ that presents a behavior closer to the ideal dynamic selection technique, for estimating the competence level of the base classifiers. We call the proposed system META-DES.Oracle since the optimization of the meta-classifier is based on the definition of the Oracle.


\paragraph{Overfitting Control Scheme}

\begin{figure*}[!ht]
	
	\begin{center}  	 
		\includegraphics[clip=,  width=0.75\textwidth]{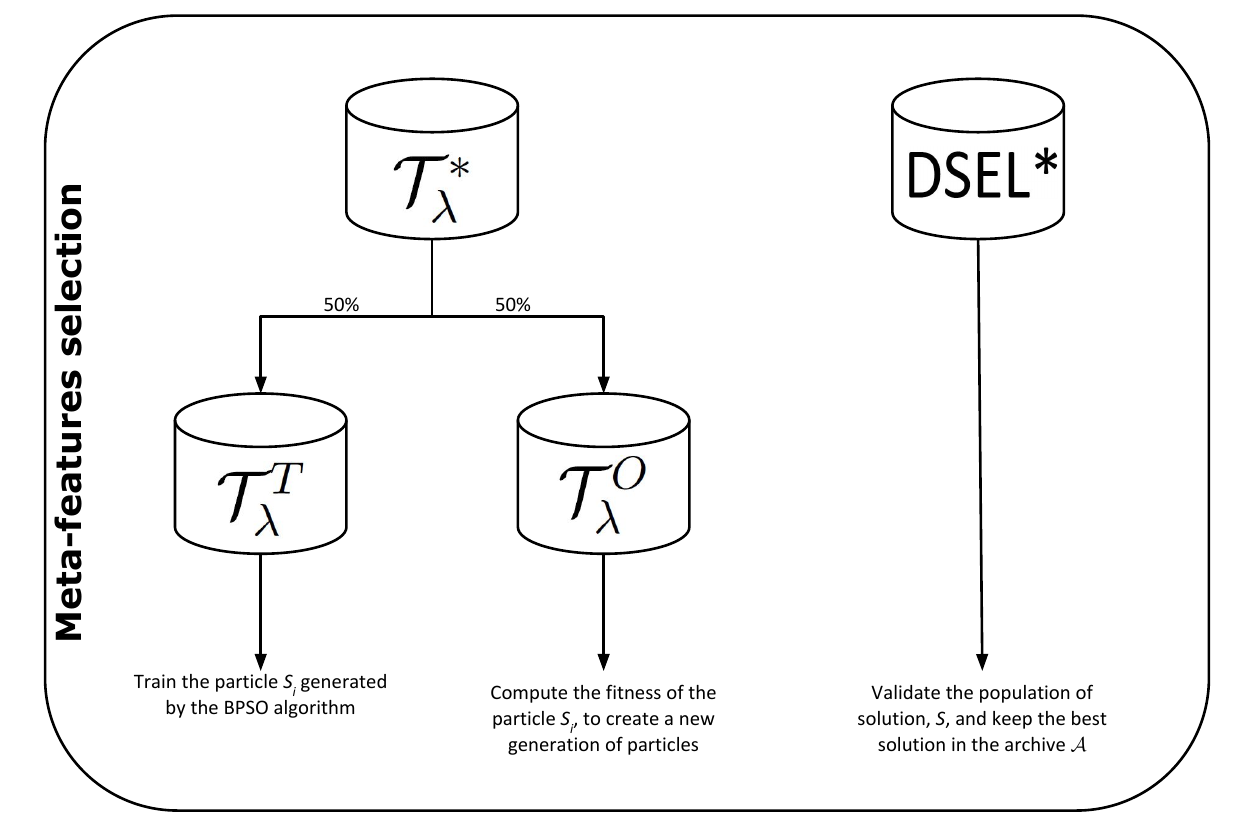}
	\end{center}
	\caption{Division of the datasets for the BPSO with global validation scheme.}
	\label{fig:globalValidation}
\end{figure*}

Since the fitness function takes into account the performance of the meta-classifier, i.e., the wrapper approach, the optimization process becomes another learning process and may be prone to overfitting~\cite{RadtkeWS06,SantosSM09,SantosOSM08}. The best solution found during the optimization routine may have overfitted the optimization dataset, and may not have a good generalization performance. To avoid overfitting, the sets used in the BPSO feature selection scheme are divided as illustrated in Figure~\ref{fig:globalValidation}. The meta-feature dataset,$\mathcal{T}_{\lambda}^{*}$, is split on the basis of 50\% for the training of the meta-classifier $\mathcal{T}_{\lambda}^{T}$ and 50\% for the optimization dataset $\mathcal{T}_{\lambda}^{O}$ which is used to guide the search in the BPSO scheme. The meta-feature vectors extracted from the dynamic selection dataset, $DSEL^{*}$, are used to validate the solutions $S_{i}$ found by the BPSO algorithm.

There are three common methods for controlling overfitting in optimization systems~\cite{SantosSM09}: Partial Validation (PV), Backwarding Validation (BV), and Global Validation (GV). In this work, we use the GV approach since previous works in the literature demonstrate that the GV is a more robust alternative for controlling overfitting in optimization techniques~\cite{RadtkeWS06,SantosSM09}. In the GV scheme (see Algorithm~\ref{alg:globalV}), at each generation, the fitness of all particles $S_{i}^{g} \in S$ are evaluated using the validation set, $DSEL^{*}$ (line 18 of the algorithm). If the fitness of the particle $S_{i}^{g}$ is better than the fitness of the particle kept in the archive, denoted by $\mathscr{A}$, $S_{i}^{g}$ is stored in the archive (lines 19 and 20). Thus, at the end of the optimization process, the particle kept in $\mathscr{A}$ is the one presenting the best fitness value, considering the validation data. The solution kept in the archive, $\mathscr{A}$, is used as the meta-classifier $\lambda$.

\begin{algorithm}[htbp]	
	\caption{BPSO meta-features selection with Global Validation}
	\label{alg:globalV}
	\begin{algorithmic}[1]
		\STATE $\mathscr{A} = \emptyset$
		\STATE Randomly initialize a swarm $S =  \left\lbrace S_{1}, \:  S_{2}, \:  ..., \:  S_{max(S)}   \right\rbrace $
		\FOR{ each generation $g \in {1, ..., max(g)}$ }
		
		\STATE "Perform all steps to generate the new solutions"
		\FOR {each particle $S_{i}^{g} \mid i = 1, \: ..., \: max(S)$}
		
		\STATE Evaluate fitness of the particle $S_{i}^{g}$ (Section~\ref{sec:fitness}).
		\IF {fitness($S_{i}^{g}$) < fitness($pBest_{i}$)}
		\STATE $pBest_{i}$ = $S_{i}^{g}$
		\ENDIF
		\IF {fitness($S_{i}^{g}$) < fitness($gBest$)}
		\STATE $gBest$ = $S_{i}^{g}$
		\ENDIF
		
		\ENDFOR	
		
		\STATE Compute the velocity of each particle using Equation~\ref{eq:velocity}.
		\STATE Update the position of each particle using Equation~\ref{eq:switching}.
		
		\FOR {each particle $S_{i}^{g+1} \mid i = 1, \: ..., \: max(S)$}
		\STATE Estimate the fitness of $S_{i}^{g+1}$ using the dataset $DSEL^{*}$.
		\IF {fitness($S_{i}^{g+1} $) < fitness($\mathscr{A}$)}
		\STATE "Store $S_{i}^{g+1} $ in the archive."
		\STATE $\mathscr{A}$ = $S_{i}^{g+1}$.
		\ENDIF
		\ENDFOR		
		\ENDFOR
		\RETURN The particle stored in the archive $\mathscr{A}$.
	\end{algorithmic}
\end{algorithm}

\subsection{Generalization Phase}
\label{sec:generalizationPhase}

\begin{algorithm}[htbp]
	\caption{Classification steps using the selector $\lambda$}
	\label{alg:generalization}
		\begin{algorithmic}[1]
			\REQUIRE Query sample $\mathbf{x}_{j,test}$
			\REQUIRE Pool of classifiers $C = \{c_{1}, \ldots, c_{M}\}$
			\REQUIRE The solution kept in the archive $\mathscr{A}$.
			\REQUIRE dynamic selection dataset $DSEL$
			\STATE $C^{'} = \emptyset$
			\FORALL{$c_{i} \in C$}
					
					\STATE Compute the meta-features selected in the archive $\mathscr{A}$ to obtain the meta-feature vector $v_{i,j}$
					\STATE input $v_{i,j}$ to $\lambda$
					
					\STATE Estimate the level of competence $\delta_{i,j}$.
				
					\STATE ``If the base classifier $c_{i}$ attain a certain level of competence it is selected to compose the ensemble $C'$.''
					\IF{ $\delta_{i,j} \ge \Upsilon$ }

						\STATE $C^{'} = C^{'} \cup \left\lbrace  c_{i} \right\rbrace$
						\STATE $\delta_{i,j}^{'} = \delta_{i,j}^{'} \cup \left\lbrace \delta_{i,j} \right\rbrace$
					\ENDIF

			\ENDFOR
			\STATE ``Each selected base classifier $c_{i,j}$ is weighted by it's competence level $\delta_{i,j}$ provided by the meta-classifier $\lambda$.'' 
			\STATE $w_{l} = WeightedMajorityVote(\mathbf{x}_{j,test},C^{'},\delta_{i,j}^{'})$
			\RETURN The predicted label $w_{l}$ for the sample $\mathbf{x}_{j,test}$
		\end{algorithmic}
\end{algorithm}

The generalization procedure is formalized by Algorithm~\ref{alg:generalization}. Given the query sample, $\mathbf{x}_{j,test}$, the region of competence $\theta_{j}$ is computed using the samples from the dynamic selection dataset $DSEL$. Following that, the output profiles, $\tilde{\mathbf{x}}_{j,test}$ of the test sample, $\mathbf{x}_{j,test}$, are calculated. The set with $K_{p}$ similar output profiles, $\phi_{j}$, of the query sample $\mathbf{x}_{j,test}$, is obtained through the Euclidean distance applied over the output profiles of the dynamic selection dataset.

For each base classifier, $c_{i}$, belonging to the pool of classifiers $C$, the meta-feature extraction process is called (Section~\ref{sec:classifierCompetence}), returning the meta-features vector $v_{i,j}$ (lines 5 and 6). Only the selected meta-features, which are kept in the archive $\mathscr{A}$ are extracted. Then, $v_{i,j}$ is used as input to the meta-classifier $\lambda$. The support, $\delta_{i,j}$, obtained by $\lambda$ for the ``competent'' meta-class, is computed as the level of competence of the base classifier, $c_{i}$, for the classification of the test sample, $\mathbf{x}_{j,test}$. The classification of the query sample, $\mathbf{x}_{j,test}$, is performed using a hybrid dynamic selection and weighting approach. First, the base classifiers that achieve a level of competence, $\delta_{i,j} > \Upsilon = 0.5$, are considered competent, and are selected to compose the ensemble, $C'$ (lines 7 to 9). Next, the decision of each selected base classifier is weighted by its level of competence, $\delta_{i,j}$, using a weighted majority voting scheme (line 13) to predict the label $w_{l}$ of the query sample $\mathbf{x}_{j,test}$. Thus, the base classifiers that attained a higher level of competence, $\delta_{i,j}$, have more influence in the final decision.

\section{Meta-Feature Extraction}
\label{sec:classifierCompetence}

A total of 15 sets of meta-features are considered, with ten sets proposed in this paper, and five coming from our previous work~\cite{CruzPR}. Each set $f_{i}$ captures a different property of the behavior of the base classifier, and can be seen as a different criterion to dynamically estimate the level of competence of the base classifier, such as the classification performance estimated in a local region of the feature space and the classifier confidence for the classification of the input sample. Using 15 distinct sets of meta-features, even though one criterion might fail due to imprecisions in the local regions of the feature space or due to low confidence results, the system can still achieve a good performance, as other meta-features are considered by the selection scheme. 

Table~\ref{table:metafeatures} shows the criterion used by each $f_{i}$, the object used to extract the meta-feature (e.g., the region of competence, $\theta_{j}$), and its categorization based on the DES taxonomy suggested in~\cite{Alceu2014}. Each set of meta-features may generate more than one feature. The size of the feature vector, $v_{i,j}$,  is $(K \times 8) + K_{p} + 6$.

\begin{table}[htbp]
    \centering
    \caption{A summary of each set of meta-features. They are categorized into the subgroups proposed in~\cite{Alceu2014}. $K$ is the size of the region of competence, $\theta_{j}$, and $K_{p}$ the size of the output profiles set $\phi_{j}$ containing the $K_{p}$ most similar output profiles of the query sample $\mathbf{x}_{j}$. The size of the meta-feature vector is  $(K \times 8) + K_{p} + 6$. The sets of meta-features marked with an * correspond to sets previously defined in~\cite{CruzPR}.} 
     \label{table:metafeatures} 
          \resizebox{1.0\textwidth}{!}{
     \begin{tabular}{|l | l | l| l| l|}
    \hline
       \textbf{Meta-Feature} & \textbf{Criterion} & \textbf{Domain} & \textbf{Object} & \textbf{No. of Features}  \\
        \hline
        $f_{Hard}$*  & Classification of the K-Nearest Neighbors & Accuracy  & $\theta_{j}$ & $K$   \\
       \hline
       $f_{Prob}$*  & Posterior probability obtained for the K-Nearest Neighbors  & Probabilistic & $\theta_{j}$ & $K$ \\
 		\hline
        $f_{Overall}$*  & Overall accuracy in the region of competence  & Accuracy & $\theta_{j}$ & 1 \\
       \hline
       $f_{Cond}$  & Conditional accuracy in the region of competence  & Accuracy & $\theta_{j}$ & 1 \\
       \hline
         $f_{Conf}$*  & Degree of confidence for the input sample  & Confidence & $\mathbf{x}_{j}$ & 1 \\
       \hline
       $f_{Amb}$  & Ambiguity in the vector of class supports  & Ambiguity & $\mathbf{x}_{j}$ & 1 \\
 		\hline

       $f_{Log}$ & 	Logarithmic difference between the class supports	&  Probabilistic & $S(\mathbf{x}_{j})$ & $K$ \\
       \hline
       $f_{PRC}$  & Probability of Random Classifier						& Probabilistic & $S(\mathbf{x}_{j})$ & $K$  \\
       \hline
       $f_{MD}$  & Minimum difference between the predictions 	& Probabilistic & $S(\mathbf{x}_{j})$ & $K$ \\
       \hline
       $f_{Ent}$  & Entropy in the vector of class supports 	& Probabilistic & $S(\mathbf{x}_{j})$ & $K$ \\
       \hline
       $f_{Exp}$	& Exponential difference between the class supports		& Probabilistic & $S(\mathbf{x}_{j})$ & $K$ \\ 
       \hline
       $f_{KL}$ & Kullback-Leibler divergence  	& Probabilistic &  $S(\mathbf{x}_{j})$ & $K$ \\
       \hline

        $f_{OP}$*  & Output profiles classification  & Behavior & $\phi_{j}$ & $K_{p}$  \\

       \hline
       $f_{Rank}$  & Classifier ranking in the feature space 	& Ranking & DSEL & 1 \\
       \hline
       $f_{Rank_{OP}}$  & Classifier ranking in the decision space 	& Behavior and Ranking & $\phi_{j}$ & 1 \\
       \hline
    \end{tabular} }
\end{table}

Given a new sample, $\mathbf{x}_{j}$, the first step in extracting the meta-features involves computing its region of competence, denoted by $\theta_{j} = \left \{ \mathbf{x}_{1}, \ldots, \mathbf{x}_{K} \right \}$. The region of competence is defined in the dynamic selection dataset $DSEL$ set using the K-Nearest Neighbor algorithm. Then, $\mathbf{x}_{j}$ is transformed into an output profile $\tilde{\mathbf{x}}_{j}$. The output profile of the instance $\mathbf{x}_{j}$ is denoted by $\tilde{\mathbf{x}}_{j} = \left\lbrace \tilde{\mathbf{x}}_{j,1}, \tilde{\mathbf{x}}_{j,2}, \ldots, \tilde{\mathbf{x}}_{j,M} \right\rbrace $, where each $\tilde{\mathbf{x}}_{j,i}$ is the decision yielded by the base classifier $c_{i}$ for the sample $\mathbf{x}_{j}$~\cite{paulo2}. Then, the similarity between $\tilde{\mathbf{x}}_{j}$ and the output profiles of the samples in $DSEL$ is obtained through the Euclidean distance. The $K_{p}$ most similar output profiles are selected to form the set $\phi_{j} = \left \{ \tilde{\mathbf{x}}_{1}, \ldots, \tilde{\mathbf{x}}_{K_{p}} \right \}$, where each output profile $\tilde{\mathbf{x}}_{k}$ is associated with a label $w_{l,k}$. 

\subsection{Local Accuracy Meta-Features}

These meta-features are based on the performance of the base classifier in a local region of the feature space surrounding the query instance $\mathbf{x}_{j}$. Three sets of meta-features using local accuracy estimation are considered:

\subsubsection{Overall Local accuracy: $f_{Overall}$}

The accuracy of $c_{i}$ over the whole region of competence $\theta_{j}$ is computed and encoded as $f_{Overall}$ (Equation~\ref{eq:ola}). 

\begin{equation}
\label{eq:ola}
f_{Overall} = \sum_{k = 1}^{K}P(w_{l} \mid \mathbf{x}_{k} \in w_{l}, c_{i} )
\end{equation}

\subsubsection{Conditional Local Accuracy: $f_{cond}$}

The local accuracy of $c_{i}$ is estimated with respect to the output classes; $w_{l}$ ($w_{l}$ is the class assigned for $\mathbf{x}_{j}$ by $c_{i}$) for the samples belonging to the region of competence, $\theta_{j}$ (Equation~\ref{eq:lca}). 

\begin{equation}
\label{eq:lca}
f_{cond} = \frac{\sum_{\mathbf{x}_{k} \in w_{l}}P(w_{l} \mid \mathbf{x}_{k}, c_{i} )}{\sum_{k = 1}^{K}P(w_{l} \mid \mathbf{x}_{k}, c_{i} )}    
\end{equation}

\subsubsection{Neighbors' hard classificationL: $f_{Hard}$}

First, a vector with $K$ elements is created. For each instance $\mathbf{x}_{k}$, belonging to the region of competence $\theta_{j}$, if $c_{i}$ correctly classifies $\mathbf{x}_{k}$, the $k$-th position of the vector is set to 1, otherwise it is 0. Thus, $K$ meta-features are computed.

\subsection{Ambiguity} 

Ambiguity measures the level of confidence the base classifier $c_{i}$ has in its answer. A common concept used to estimate the confidence of a classifier is based on the margin theory~\cite{boosting,Breiman:1999}. The margin of a classifier is regarded as a good indicator of the classifier's confidence.
Two meta-features are considered: one based on the maximum margin theory $f_{conf}$ and one based on the minimum margin theory $f_{amb}$. Since these meta-features do not take into account the correct label of the sample, they are extracted directly from the query sample, $\mathbf{x}_{j}$. 

\subsubsection{Classifier's confidence: $f_{Conf}$}

The perpendicular distance between the input sample, $\mathbf{x}_{j}$, and the decision boundary of the base classifier $c_{i}$ is calculated and encoded as $f_{conf}$. The value of $f_{conf}$ is normalized to a $[0-1]$ range using the Min-max normalization. 

\subsubsection{Ambiguity: $f_{Amb}$}

This information is simply computed by the difference between scores of the class with highest support and the second highest one for the query sample, $\mathbf{x}_{j}$, e.g., consider that for a 3-class classification problem, the scores obtained by the base classifier $c_{i}$ for a given query sample, $\mathbf{x}_{j}$, are $0.65$, $0.30$ and $0.05$. Then, the ambiguity value is $f_{amb} = 0.65 - 0.30 = 0.35$. A higher value in $f_{amb}$ means that the classifier decision is less ambiguous.

\subsection{Probabilistic Meta-Features}
 
This class of meta-features is based on probabilistic models that are applied over the vector of class supports produced by the base classifier $c_{i}$ for the classification of a given query sample. The motivation behind probabilistic measures derives from the observation that classifiers that perform worse than the random classifier, i.e., a classifier that randomly select the classes with equal probabilities, deteriorate the majority voting performance. In contrast, if the base classifiers are significantly better than the random classifier, they are likely to improve the majority voting accuracy~\cite{WoloszynskiKPS12}. Hence, each set of meta-features in this group estimates the probability that the performance of a given base classifier $c_{i}$ is significantly different from that of a random classifier derived from different probabilistic and information theory perspectives~\cite{Woloszynski,WoloszynskiK10,WoloszynskiKPS12,zbMATH05935973,WoloszynskiK09}. 

%

For the definitions below, let $S(\mathbf{x}_{k}) = \left\lbrace S_{1}(\mathbf{x}_{k}), \ldots, S_{L}(\mathbf{x}_{k}) \right\rbrace$  be the vector of class supports estimated by the base classifier $c_{i}$ for a given sample, $\mathbf{x}_{k}$, where each value $S_{l}(\mathbf{x}_{k}), l = 1, 2 \; \ldots, \; L$  represents the support given to the $l$-th class and $\sum\limits_{l = 1}^{L} S_{l}(\mathbf{x}_{k}) = 1$. Let $S_{lk}(\mathbf{x}_{j})$ be the support given by the base classifier $c_{i}$ for the correct class label of $\mathbf{x}_{j}$. The output of the random classifier follows a uniform distribution, and is denoted by $RC = \left \lbrace \frac{1}{L}, ... , \frac{1}{L} \right \rbrace$.

\subsubsection{Posterior probability: $f_{Prob}$ } First, a vector with $K$ elements is created. Then, for each instance $\mathbf{x}_{k}$, belonging to the region of competence $\theta_{j}$, the posterior probability of $c_{i}$, $P(w_{l}\mid \mathbf{x}_{k})$ is computed and inserted into the $k$-th position of the vector. Consequently, $K$ meta-features are computed. 

\subsubsection{Logarithmic: $f_{Log}$}

First, a vector with $K$ elements is created, $f_{Log} = \left \{ f_{Log}(1), \; \:  ..., \; f_{Log}(K) \right \}$. For each instance, $\mathbf{x}_{k}$, belonging to the region of competence $\theta_{j}$, the support obtained by the base classifier $c_{i}$ for the correct class label, $S_{lk}(\mathbf{x}_{k})$, is estimated. Then, a logarithmic function~\cite{WoloszynskiK09} is applied to $S_{lk}(\mathbf{x}_{k})$ (Equation~\ref{eq:log}). The function is used such that the value of the meta-feature is negative if the support obtained for the correct class label is lower than the support obtained from random guessing (i.e., $S_{lk}(\mathbf{x}_{j}) < \frac{1}{L}$) and positive otherwise. The result of the logarithmic function is inserted into the $k$-th position of the vector. Hence, $K$ meta-features are computed. 

\begin{equation}
\label{eq:log}
f_{log}(k) = 2 \times S_{lk}(\mathbf{x}_{k})^{ \frac{log(2)}{log(L)}} - 1 
\end{equation}

\subsubsection{Entropy: $f_{Ent}$}

The entropy measures the degree of uncertainty in the vector of supports, $S(\mathbf{x}_{k})$, obtained by the base classifier, $c_{i}$. The meta-feature is calculated as follows: first, a vector with $K$ elements is created, $f_{Ent} = \left \{ f_{Ent}(1), \; \:  ..., \; f_{Ent}(K) \right \}$. Then, for each instance, $\mathbf{x}_{k}$, belonging to the region of competence, $\theta_{j}$, the entropy of the vector of class supports is computed, and inserted in the $k$-th position of the vector $f_{Ent}$ (Equation~\ref{eq:entropy}). Thus, $K$ meta-features are computed. 

\begin{equation}
\label{eq:entropy}
f_{Ent}(k) = - \sum_{l=1}^{L} S_{l}(\mathbf{x}_{k}) log(S_{l}(\mathbf{x}_{k})) 
\end{equation}

\subsubsection{Minimal difference: $f_{MD}$}

First, a vector with $K$ elements is created, $f_{MD} = \left \{ f_{MD}(1), \; \:  ..., \; f_{MD}(K) \right \}$. Then, for each  sample, $\mathbf{x}_{k}$, belonging to the region of competence, $\theta_{j}$, the Minimal Difference (as proposed in~\cite{zbMATH05935973}) is computed as the difference between the support obtained by the base classifier $c_{i}$ for the correct class label of $\mathbf{x}_{k}$, $S_{lk}(\mathbf{x}_{k})$, and those obtained by $c_{i}$ for each of the other classes, $S_{l}(\mathbf{x}_{k}) \mid \; l \neq lk$, are calculated. The difference which produces the minimal value is inserted in the $k$-th position of the vector $f_{MD}$ (Equation~\ref{eq:md}). Thus, $K$ meta-features are computed.

\begin{equation}
\label{eq:md}
f_{MD}(k) = min_{l \; \in \; L, \; l \; \neq \; lk} \left[ S_{l}(\mathbf{x}_{k}) - S_{lk}(\mathbf{x}_{k}) \right]  
\end{equation}

\subsubsection{Kullback-Leibler Divergence: $f_{KL}$}

The Kullback-Leibler (KL) divergence~\cite{Kullback51klDivergence} estimates the competence of a base classifier $c_{i}$ from the information theory perspective~\cite{WoloszynskiKPS12}. The meta-feature is computed as follows: first, a vector with $K$ elements is created, $f_{KL} = \left \{ f_{KL}(1), \; \:  ..., \; f_{KL}(K) \right \}$. Then, for each member, $\mathbf{x}_{k}$, belonging to the region of competence $\theta_{j}$, the KL divergence between the vector of class supports, $S(\mathbf{x}_{k}) = \left\lbrace S_{1}(\mathbf{x}_{k}), \ldots, S_{L}(\mathbf{x}_{k}) \right\rbrace$, obtained by the base classifier, $c_{i}$, and those obtained by the random classifier, $RC = \left \lbrace \frac{1}{L}, ... , \frac{1}{L} \right \rbrace$ is computed. The result of the KL divergence is inserted in the $k$-th position of the vector $f_{KL}$ (Equation~\ref{eq:KL}). Consequently, $K$ meta-features are calculated.

\begin{equation}
\label{eq:KL}
f_{KL}(k) = \sum_{l = 1}^{L} S_{l}({\mathbf{x}}_{k}) log \frac{S_{l}({\mathbf{x}}_{k})}{RC}
\end{equation}

\subsubsection{Exponential: $f_{Exp}$}

First, a vector with $K$ elements is created, $f_{Exp} = \left \{ f_{Exp}(1), \; \:  ..., \; f_{Exp}(K) \right \}$. For each sample, $\mathbf{x}_{k}$, belonging to the region of competence $\theta_{j}$, the support obtained by the base classifier $c_{i}$ for the correct class label, $S_{lk}(\mathbf{x}_{k})$, is estimated. Next, an exponential function~\cite{WoloszynskiK09} is applied over $S_{lk}(\mathbf{x}_{k})$ to compute $f_{Exp}$ (Equation~\ref{eq:exponential}). Using the exponential function, the value of $f_{Exp}$ increases exponentially when the value of $S_{lk}(\mathbf{x}_{k})$ is higher than that obtained from random guessing ($S_{lk}(\mathbf{x}_{k}) > \frac{1}{L}$), and is negative otherwise. The result of the exponential function is inserted in the $k$-th position of the vector. Hence, $K$ meta-features are computed.

\begin{equation}
\label{eq:exponential}
f_{Exp}(k) = 1 - 2^{-1 \frac{(L-1) S_{lk}({x}_{k})} {1 - S_{lk}({x}_{k})} } 
\end{equation}

\subsubsection{Randomized Reference Classifier: $f_{PRC}$}

First, a vector with $K$ elements is created, $f_{PRC} = \left \{ f_{PRC}(1), \; \:  ..., \; f_{PRC}(K) \right \}$. For each sample, $\mathbf{x}_{k}$, belonging to the region of competence $\theta_{j}$, the conditional probability of correct classification estimated by the randomized reference classifier (RRC) proposed in~\cite{Woloszynski}\footnote{Matlab code for this technique is available on:  \url{http://www.mathworks.com/matlabcentral/fileexchange/28391-a-probabilistic-model-of-classifier-competence} }. The result is inserted in the $k$-th position of the vector. Thus, $K$ meta-features are computed.


\subsection{Behavior meta-features}

These measures take into consideration information extracted from the decision space, i.e., the outputs or behavior of the classifiers in the pool, rather than information from the feature space. Global information about the whole pool of classifiers is considered. Furthermore, many authors have successfully utilized DES criteria based on classifier behavior in estimating the competence of base classifiers~\cite{paulo2,logid,CruzPR}.

\subsubsection{Output profiles classification: $f_{OP}$}
	
First, a vector with $K_{p}$ elements is created. Then, for each member, $\tilde{\mathbf{x}}_{k}$, belonging to the set of output profiles, $\phi_{j}$, if the label produced by $c_{i}$ for $\mathbf{x}_{k}$ is equal to the label $w_{l,k}$ of $\tilde{\mathbf{x}}_{k}$, the $k$-th position of the vector is set to 1, otherwise it is 0. A total of $K_{p}$ meta-features are extracted.

\subsection{Ranking Meta-Features}

Ranking methods for estimating the competence of base classifiers have been proposed in~\cite{classrank}. The ranking is computed such that classifiers with higher ranking values are more likely to be competent. In this work, we consider two types of ranking meta-features, one based on the feature space, and the other on the decision space. They are defined below:

\subsubsection{Simplified classifier rank: $f_{Rank}$}

This meta-feature is inspired by the simplified classifier rank technique proposed in~\cite{lca}. The first step in extracting the ranking meta-feature is to order the instances in $DSEL$ by its distance to the query sample $\mathbf{x}_{j}$. $f_{Rank}$ is computed as the number of consecutive correct predictions made by the base classifier $c_{i}$, starting from the closest sample to  $\mathbf{x}_{j}$. The search stops when the first misclassification is made. 

\subsubsection{classifier rank OP: $f_{Rank_{OP}}$}

This meta-feature is computed similarly to the previous $f_{rank}$. However the search is conducted in the decision space, using the output profiles, $\phi_{j}$, rather than the dataset $DSEL$. Hence, the first step is to order the output profiles in $\phi_{j}$ by their similarity to the output profile of the query sample $\tilde{\mathbf{x}}_{j}$. Then, the number of consecutive correct predictions made by the base classifier $c_{i}$ is computed as $f_{Rank_{OP}}$.

\section{Case study using synthetic data}
\label{sec:synthetic}

\begin{figure}
   \begin{center}  	 
       	  \includegraphics[clip=, width=0.45\textwidth]{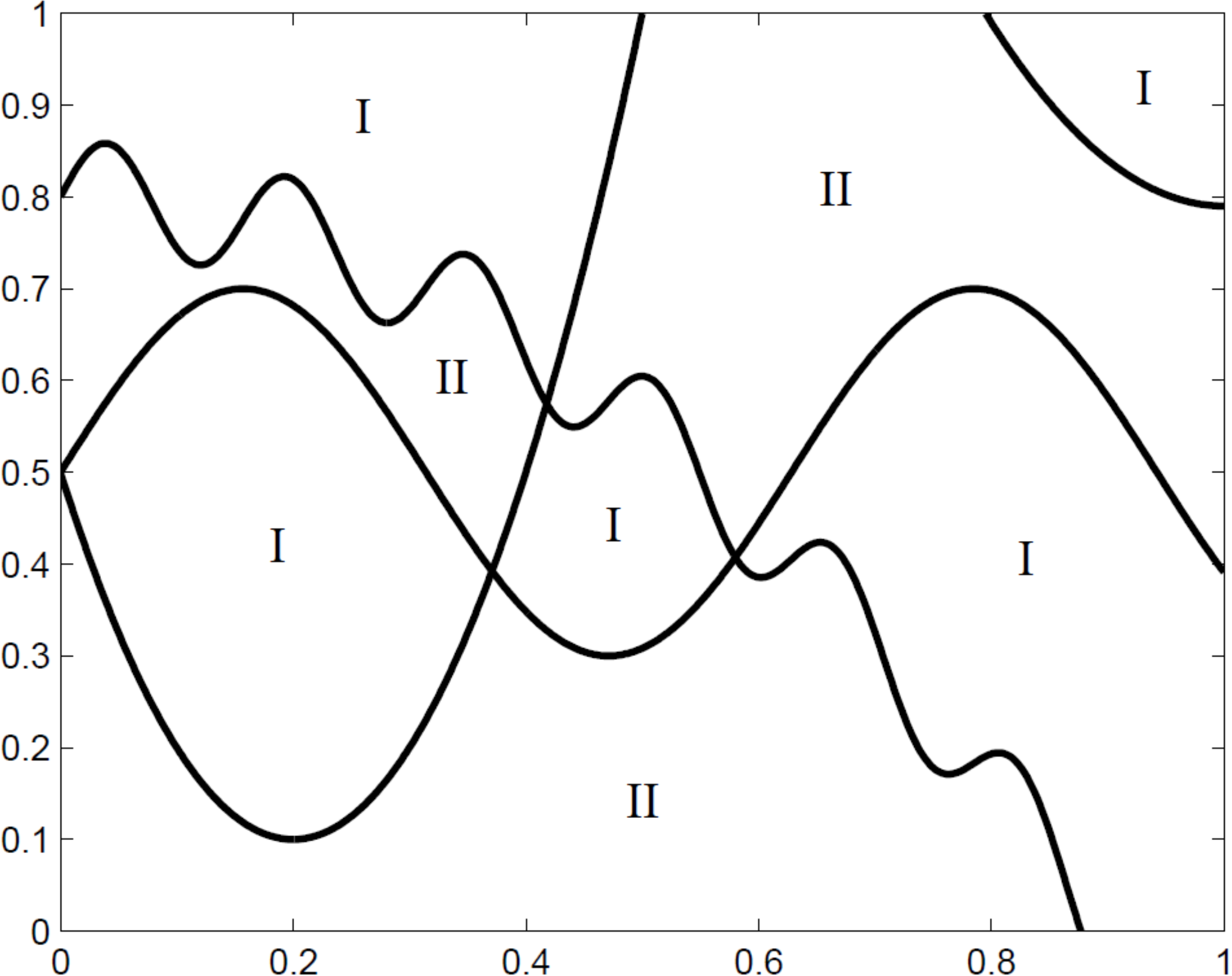}
   \end{center}
   \caption{The P2 Problem. The symbols I and II represent the area of the classes, 1 and 2, respectively}

\label{fig:P2Problem}	  
\end{figure}

In this section, we conduct experiments using a synthetic dataset in order to illustrate the benefits of the meta-feature selection process and compare different versions of the META-DES framework for solving a problem with a complex non-linear geometry using a pool composed of linear classifiers. The P2 is a two-class problem, presented by Valentini~\cite{Valentini05}, in which each class is defined in multiple decision regions delimited by polynomial and trigonometric functions (Equations~\ref{eq:problem1},~\ref{eq:problem2},~\ref{eq:problem3} and~\ref{eq:problem4}). As in~\cite{henniges}, $E4$ was modified such that the area of each class is approximately equal. The P2 problem is illustrated in Figure~\ref{fig:P2Problem}. It is impossible to solve this problem using a single linear classifier, and the performance of the best possible linear classifier is around 50\%.

\begin{eqnarray} 
\label{eq:problem1}
E1(x) = sin(x) + 5 \\
\label{eq:problem2}
E2(x) = (x - 2)^{2} + 1 \\
\label{eq:problem3}
E3(x) = -0.1 \cdot x^{2} + 0.6sin(4x) + 8 \\
\label{eq:problem4}
E4(x) = \frac{(x - 10)^{2}}{2} + 7.902 
\end{eqnarray} 

\begin{figure}
   \begin{center}  	 
       	  \includegraphics[clip=, width=0.600\textwidth]{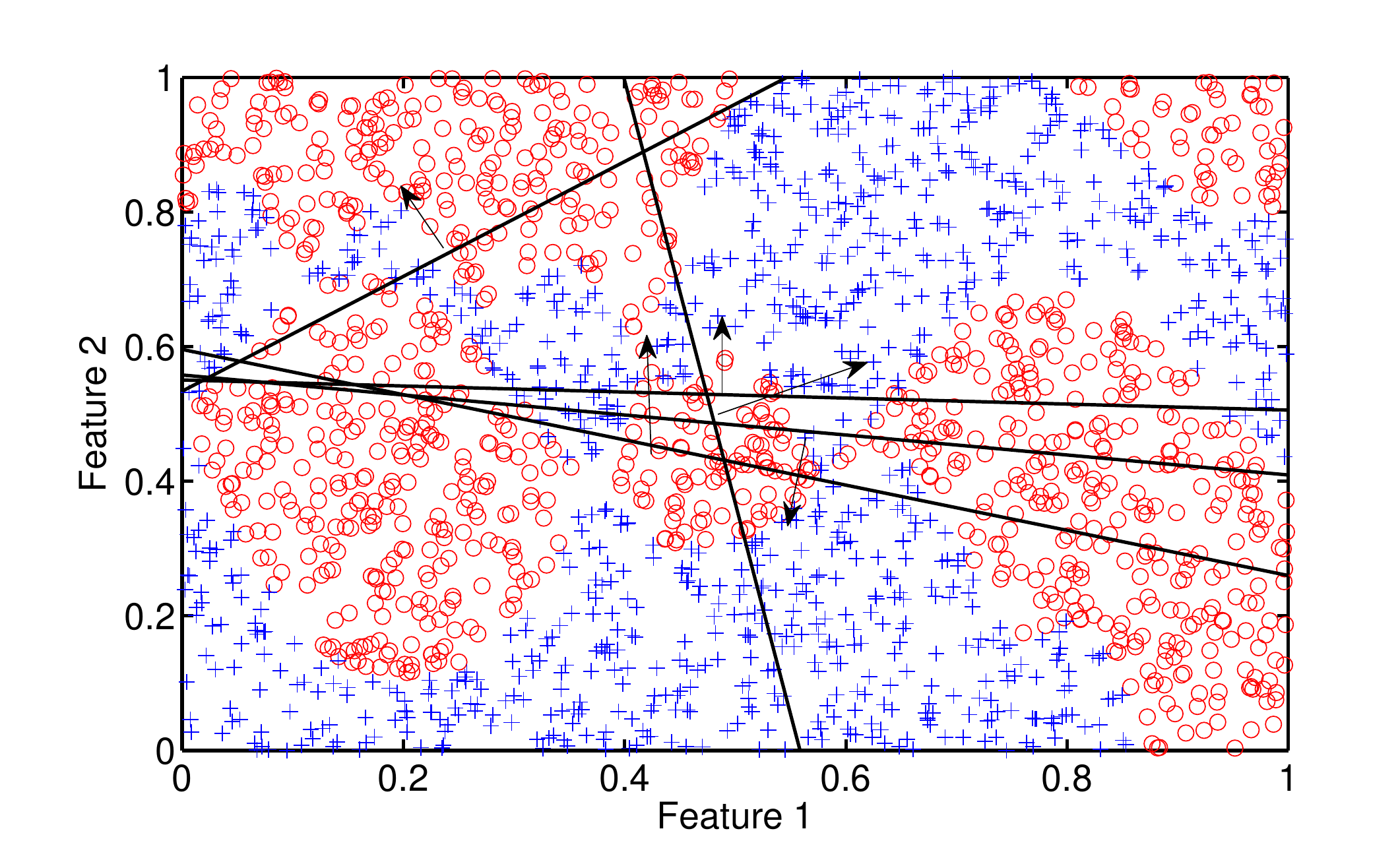}
   \end{center}
\caption{Five Perceptrons generated using the bagging technique for the P2 Problem. The arrows in each Perceptron point to the region of class 1 (red circle).}
\label{fig:FivePerceptrons}	  
\end{figure}

For this illustrative example, the P2 problem was generated as in~\cite{reportarXiv}: 500 samples for training ($\mathcal{T}$), 500 instances for the meta-training dataset ($\mathcal{T}_{\lambda}$), 500 instances for the dynamic selection dataset $DSEL$, and 2000 samples for the test set, $\mathcal{G}$. The pool of classifiers is composed of 5 Perceptrons (shown in Figure~\ref{fig:FivePerceptrons}). The best classifier of the pool (Single Best) achieves an accuracy of 53.5\%. The performance of all other base classifiers is around the 50\% mark. The Oracle result of this pool obtained a recognition performance of 99.5\%. In other words, there is at least one base classifier that predicts the correct label for 99.5\% of the test instances. The problem lies in selecting the competent classifiers in order to achieve a classification accuracy close to the Oracle.

\begin{figure}
	\centering
	\subfigure[META-DES]{\includegraphics[width=3.1in,clip=]{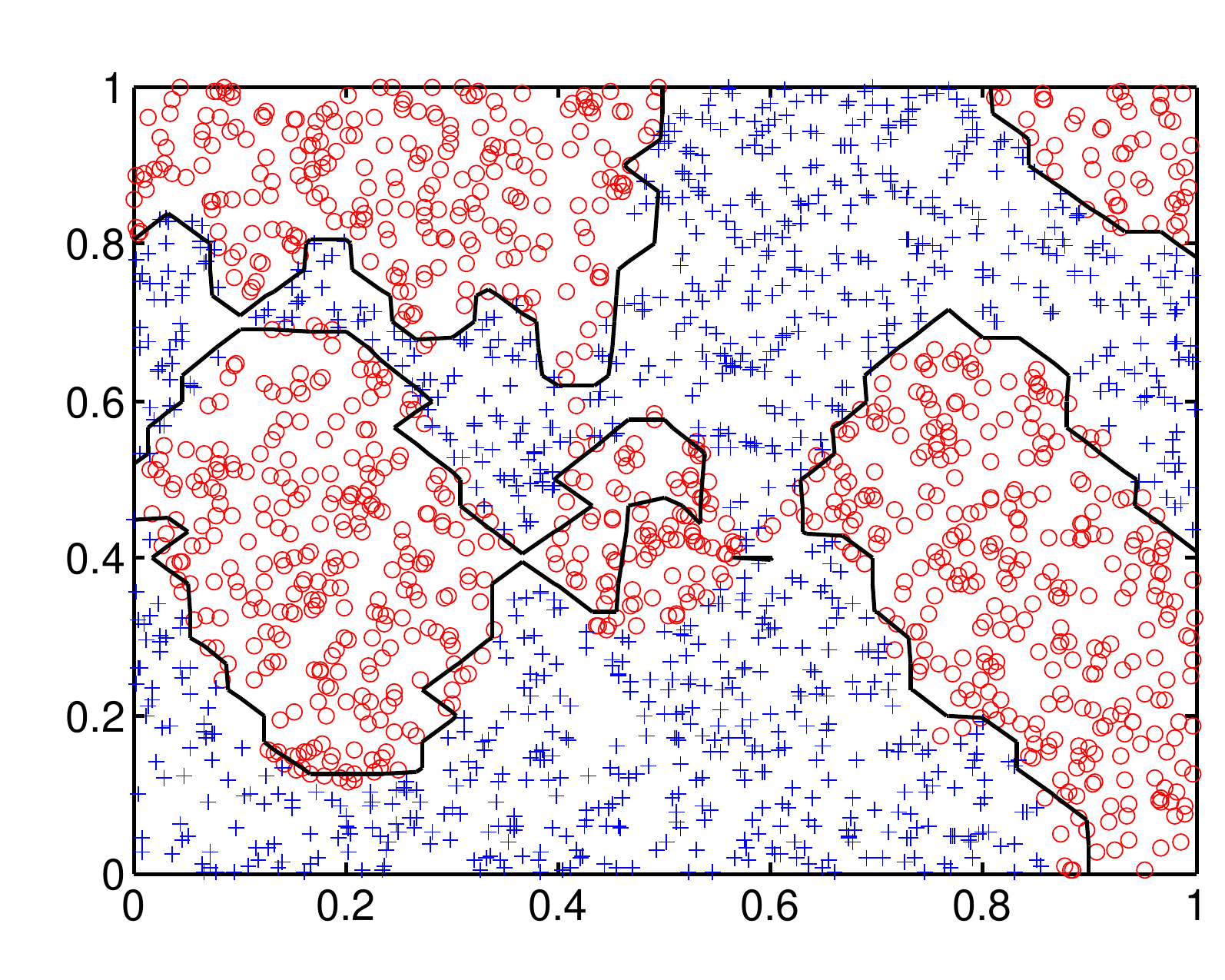} } 
	\subfigure[META-DES.Oracle]{\includegraphics[width=3.1in,clip=]{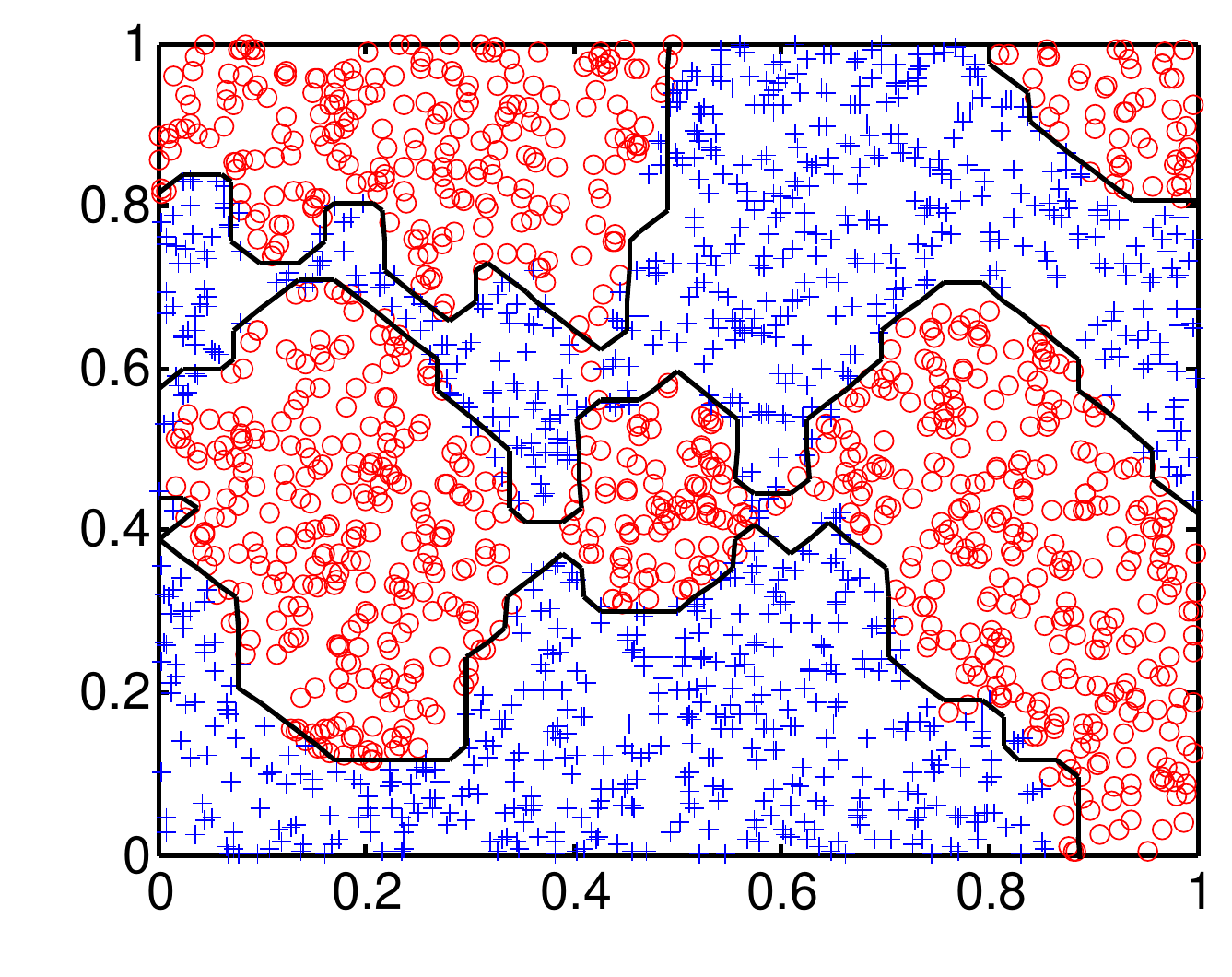} } 
	\caption{Decision boundary obtained by two versions of the META-DES framework. (a) Original META-DES (b) The proposed META-DES.Oracle}
	\label{fig:metadesresults}	  
\end{figure}

Figures~\ref{fig:metadesresults} (a) and (b) show the decision boundary obtained by the META-DES~\cite{CruzPR}, and the proposed META-DES.Oracle, respectively\footnote{The results achieved by different dynamic and static ensemble techniques for the P2 problem are presented in the following report~\cite{reportarXiv}.}. We can observe that the META-DES.Oracle obtains a really good approximation of the real decision boundary for the P2 problem. The META-DES.Oracle proposed in this paper obtained a recognition accuracy of 97\%, while the accuracy of the META-DES was 94.5\%~\cite{reportarXiv}. Using the extended sets of meta-features and the meta-feature selection procedure based on the Oracle definition, we observed a significant gain in performance for the P2 problem. Thus, it is possible to reduce the big gap that exists between the performances of the current state-of-the-art DES techniques and the ideal one, the Oracle.

\section{Experiments}
\label{sec:experiments}

\subsection{Datasets}

The experiments were conducted on the same 30 classification datasets used in our previous work~\cite{CruzPR,ijcnn2015}. The key features of each dataset are shown in Table~\ref{table:datasets}. 

\begin{table}[htbp]
	\centering
	\caption{Dataset considered in this work and their main features. They are presented in alphabetical order.}
	\label{table:datasets} 
	\resizebox{0.65\textwidth}{!}{
		\begin{tabular}{|c| c| c| c| c|}
			\hline
			\textbf{Database} & \textbf{\# samples} & \textbf{\# features} & \textbf{\# Classes} & \textbf{Repository}\\
			\hline
			
			\textbf{Adult} & 48842 & 14 & 2 & UCI  \\        
			\hline
			\textbf{Banana}  & 1000 & 2 &	2 &  PRTOOLS  \\
			\hline
			\textbf{Blood transfusion} & 748 & 4 &	2 &  UCI  \\
			\hline 
			\textbf{Breast (WDBC)} & 568 & 30 & 2 &  UCI \\
			\hline
			\textbf{Cardiotocography (CTG)} & 2126 & 21 & 3 &  UCI \\    
			\hline
			\textbf{Ecoli} & 336 & 7 & 8 &  UCI  \\    
			\hline
			\textbf{Steel Plate Faults} & 1941 & 27 & 7 &  UCI \\   
			\hline
			\textbf{Glass} & 214 & 9 & 6  &  UCI  \\                       
			\hline
			\textbf{German credit} & 1000 & 20 &2  &  STATLOG \\
			\hline
			\textbf{Haberman's Survival} & 306 & 3 & 2 & UCI \\
			\hline
			\textbf{Heart} & 270 & 13  & 2  &  STATLOG \\
			\hline
			\textbf{ILPD} & 583 & 10 & 2  &  UCI \\                       
			\hline									
			\textbf{Ionosphere} & 315 &	34 & 2 &  UCI  \\
			\hline
			\textbf{Laryngeal1} & 213 & 16 & 2 &  LKC \\        
			\hline   
			\textbf{Laryngeal3} & 353 & 16 & 3 &  LKC \\    
			\hline
			\textbf{Lithuanian}  & 1000 & 2 & 2 &  PRTOOLS \\
			\hline 
			\textbf{Liver Disorders} & 345 & 6 & 2 & UCI  \\
			\hline
			\textbf{MAGIC Gamma Telescope}  & 19020 & 10 & 2 &  KEEL \\
			\hline  			
			\textbf{Mammographic}  & 961 & 5 & 2 &  KEEL \\
			\hline  
			\textbf{Monk2}  & 4322 & 6 & 2 &  KEEL  \\
			\hline  			
			\textbf{Phoneme} & 5404 & 6 & 2 &  ELENA  \\   
			\hline
			\textbf{Pima} & 768 & 8 & 2 & UCI  \\
			\hline		
			\textbf{Satimage} & 6435 & 19 & 7 & STATLOG \\    
			\hline 			
			\textbf{Sonar} & 208 &	60 & 2 &  UCI \\
			\hline
			\textbf{Thyroid} &  215 & 5 & 3 &  LKC \\
			\hline
			\textbf{Vehicle} & 846 & 18 & 4 &  STATLOG \\
			\hline
			\textbf{Vertebral Column} & 310 & 6 & 2 &  UCI \\          
			\hline	
			\textbf{WDG V1} & 5000 & 21 & 3 &  UCI \\    
			\hline
			\textbf{Weaning} & 302 & 17 & 2 &  LKC \\
			\hline
			\textbf{Wine} & 178 & 13 & 3 &  UCI \\
			\hline   
			
		\end{tabular}
	}
\end{table} 

\subsection{Experimental protocol}

For each dataset, the experiments were conducted using 20 replications. For each replication, the datasets were divided using the holdout method~\cite{hastie_09} on the basis of 50\% for training, 25\% for the dynamic selection dataset, $DSEL$, and 25\% for the test set, $\mathcal{G}$. The divisions were performed while maintaining the prior probabilities of each class. For the proposed META-DES-Oracle, 25\% of the training data was used in the meta-training process $\mathcal{T}_{\lambda}$.

For the two-class classification problems, the pool of classifiers was composed of 100 Perceptrons generated using the Bagging technique. For the multi-class problems, the pool of classifiers was composed of 100 multi-class Perceptrons. The use of linear Perceptron classifiers was motivated by the results reported in Section~\ref{sec:synthetic} showing that the META-DES framework can solve non-linear classification problems with complex decision boundaries using only a few linear classifiers. The values of the hyper-parameters, $K$, $K_{p}$ and $h_{c}$, were set at 7, 5 and 70\%, respectively. They were selected empirically based on previous publications~\cite{ijcnn2011,icpr2014,CruzPR}. Hence, the size of the meta-feature vector is 67 ($(7 \times 8)$ $+ 5 + 6$).

The parameters of the BPSO algorithm were set based on previous work in the literature~\cite{ChuangTY11,Chuang:2008,FirpiG04}: the population size was set at 20, the maximum number of generations $max(g) = 100$. The weight function, $w = 1.0$, and acceleration coefficients, $c_{1} = c_{2} = 2.0$, were set using the standard values from~\cite{Kennedy:PSO}. Moreover, the optimization process was stopped if the fitness of the best solution $gBest$ failed to improve after 5 consecutive iterations. Since the BPSO optimization process is a stochastic algorithm, for each replication, the BPSO was run 30 times. The best result, considering the Global Validation overfitting control scheme, was used for generalization phase.

\subsection{Analysis of the selected meta-features}
\label{sec:metafeaturesanalysis}

In this section, we analyze the set of meta-features that are selected by the proposed technique. The objective of this analysis is: (1) to verify whether different sets of meta-features are better suited for different classification problems; and (2) to identify whether or not the proposed sets of meta-features are relevant. 

\begin{figure}[H]
	
	\begin{center}  	 
		\includegraphics[clip=,  width=1.00\textwidth]{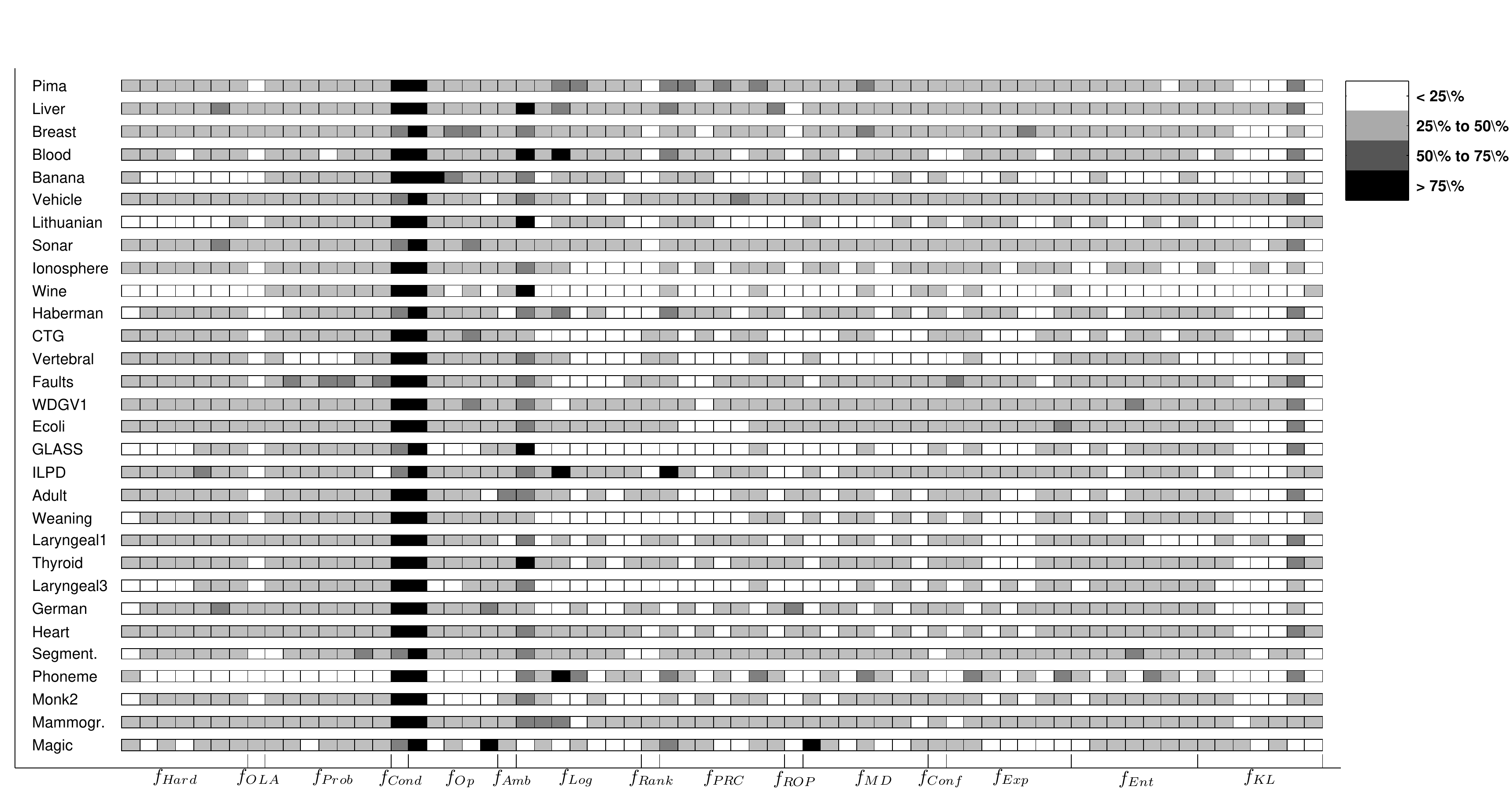}
	\end{center}
	\caption{The frequency at which each individual meta-feature is selected over 20 replications. Each dataset is evaluated separately. The color of each square represents the frequency at which each meta-feature is selected. A white square indicates that the corresponding meta-feature was selected less than 25\% of the time. A light grey square means the meta-feature was selected with a frequency between 25\% and 50\%. A dark grey square represents a frequency of 50\% to 75\%, and a black square represents a frequency of selection higher than 75\%.}
	\label{fig:selectedFeatures}
\end{figure}

In the first analysis, we compare how often each individual meta-feature was selected. Figure~\ref{fig:selectedFeatures} illustrates the selection frequency per meta-feature, considering 20 replications. We present the results for each dataset separately. Each square represents an individual meta-feature.
The color of each square represents the frequency that each meta-feature is selected. A white square indicates that the corresponding meta-feature was selected less than 25\% of the time. A light grey square means the meta-feature was selected with a frequency between 25\% and 50\%. A dark grey square represents a frequency of 50\% to 75\%, and a black square represents a frequency of selection higher than 75\%.

\begin{figure*}[htbp]
 \begin{center}  	 
       	  \includegraphics[clip=,  width=0.950\textwidth]{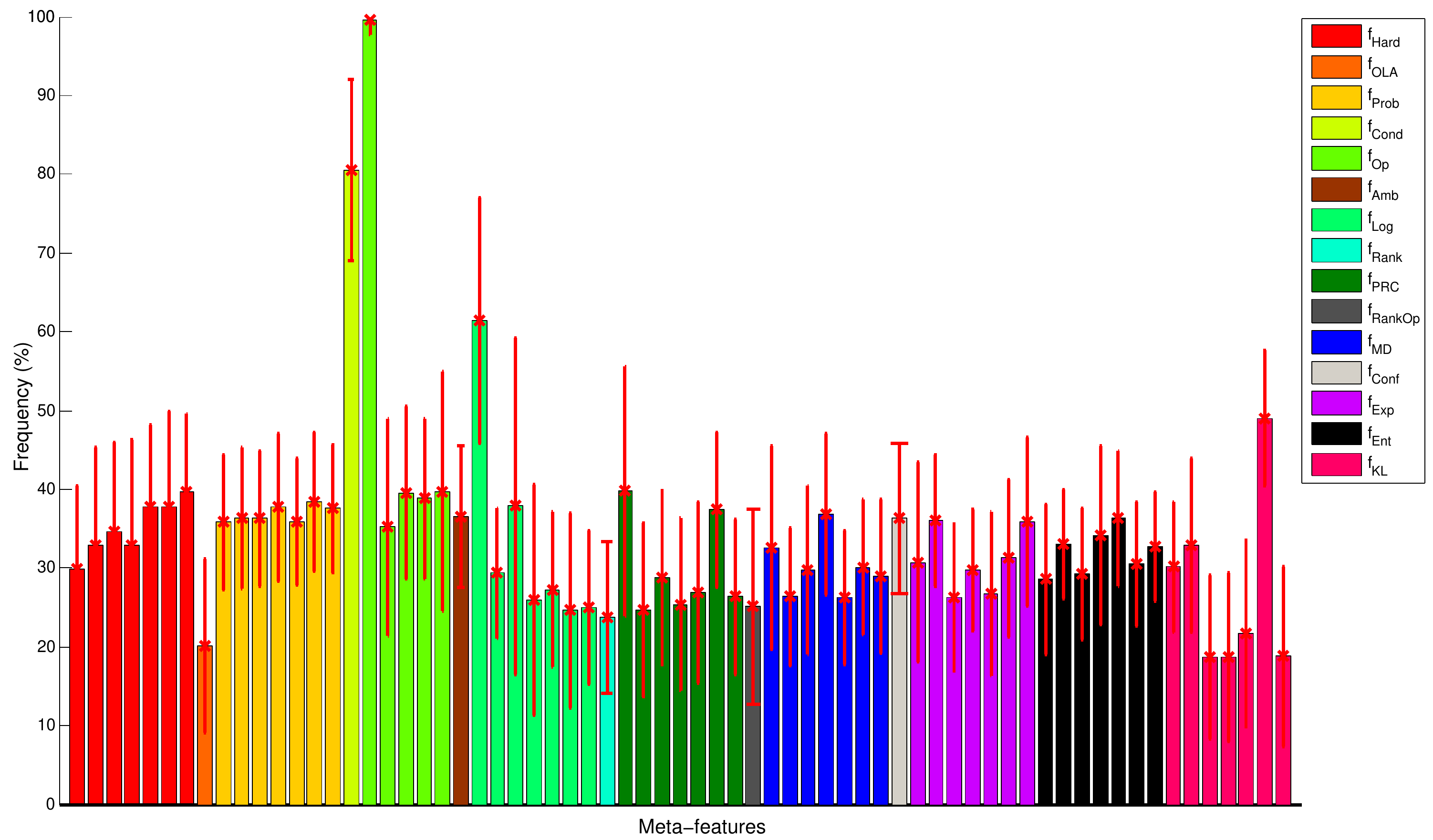}
   \end{center}
\caption{Average frequency and standard deviation per meta-feature considering 30 classification problems.}
\label{fig:selectedFeatures2}
\end{figure*}

It can be seen that the frequency at which each meta-feature is selected varies considerably between different datasets. For instance, the meta-feature based on the classification of the neighbor samples, $f_{hard}$, was selected with a frequency between 25 and 50\% in the majority of datasets. However, for the Wine dataset, it was not selected at all. 
The only exceptions are for the meta-feature sets, $f_{OP}$, which presented a 100\% frequency of selection for all 30 datasets, and $f_{cond}$. This finding demonstrates that distinct classification problems require a different set of meta-features in order to better address the behavior of the Oracle. Different problems are associated with different degrees of data complexity~\cite{HoB02}, and may require a distinct set of meta-features in order to obtain a meta-classifier that presents a behavior closer to the Oracle for estimating the competence of the base classifiers. Hence, the results show that the choice of the best set of meta-features is problem-dependent. In addition, we can see that each individual meta-feature is selected for at least 20\% of the datasets, considering all 30 classification problems (Figure~\ref{fig:selectedFeatures2}). Hence, we believe that all sets of meta-features proposed in this work are relevant.

\subsection{Comparative study}
\label{sec:comparative}

In this section, we compare the results obtained by the proposed META-DES.Oracle, which is based on 15 sets of meta-features, against the previous versions of the META-DES framework, which are based only on five sets of meta-features defined in~\cite{CruzPR}. The objective of this comparative study is to answer the following research questions: (1) Does the optimization based on the Oracle behavior lead to a significant gain in classification accuracy? (2) Does the use of more meta-features lead to a more robust DES system? 

The following versions of the META-DES framework are compared in this section: 

\begin{enumerate}
\item \textbf{S-shaped GV:} The proposed META-DES.Oracle using S-shaped transfer function with global validation.
\item \textbf{V-shaped GV:} The proposed META-DES.Oracle using V-shaped transfer function with global validation.
\item \textbf{S-Shaped:} The proposed META-DES.Oracle using S-shaped transfer function without global validation.
\item \textbf{V-Shaped:} The proposed META-DES.Oracle using V-shaped transfer function without global validation.
\item \textbf{META-DES.ALL:} The framework using the 15 sets of meta-features proposed in this work without the optimization process.
\item \textbf{META-DES.H:} The Hybrid version, META-DES.H proposed in~\cite{ijcnn2015}. 
\item \textbf{META-DES:} The first version of the META-DES framework~\cite{CruzPR}. 
\end{enumerate}

\begin{table}[h!] 
\centering 
\caption{Comparison of different versions of the META-DES framework. We present the results of statistical tests at the end of the table.} 
\label{table:resultsBPSO}  
\resizebox{1.00	\textwidth}{!}{  
\begin{tabular}{|c c c c c c || c c | }  
\hline  
 
Dataset & S-Shaped GV & V-Shaped GV & S-Shaped & V-Shaped & META-DES.ALL & META-DES.H~\cite{ijcnn2015} & META-DES~\cite{CruzPR} \\ 
 \hline

\textbf{Adult} & 87.29(2.02) & \textbf{87.74(2.04)} & 87.67(2.13) & 87.67(2.03) & 85.17(3.15) & 87.29(1.80) & 87.22(1.84) \\ 

\textbf{Banana} & 94.66(1.09) & 94.54(1.16) & 94.39(1.14) & 94.80(0.99) & \textbf{95.69(1.35)} & 94.51(2.36) & 94.42(2.37) \\ 

\textbf{Blood} & 79.44(1.84) & 79.38(1.76) & 79.79(1.38) & 79.20(1.69) & \textbf{79.91(0.79)} & 78.25(1.37) & 78.31(1.52) \\ 

\textbf{Breast} & 96.78(0.82) & 96.71(0.86) & 96.71(0.86) & 96.71(0.86) & 96.86(0.85) & 97.25(0.47) & \textbf{97.41(0.50)} \\ 

\textbf{CTG} & 86.73(1.23) & 86.37(1.10) & 86.81(1.06) & 86.68(1.16) & \textbf{87.10(0.99)} & 86.08(1.24) & 86.04(1.14) \\ 

\textbf{Ecoli} & \textbf{81.83(3.00)} & 81.57(3.47) & 81.83(3.22) & 81.44(3.63) & 78.70(3.22) & 80.66(3.48) & 80.92(3.76) \\ 

\textbf{Faults} & \textbf{69.52(0.95)} & 69.32(1.18) & 68.93(1.15) & 69.02(1.46) & 69.02(1.55) & 68.95(1.04) & 68.72(1.19) \\ 

\textbf{GLASS} & 67.09(3.89) & 66.46(4.22) & 66.88(3.71) & 66.04(4.12) & \textbf{68.77(3.71)} & 65.21(3.53) & 65.21(3.65) \\ 

\textbf{German} & 75.03(1.99) & \textbf{76.58(1.99)} & 75.43(1.92) & 76.05(1.67) & 71.56(2.91) & 74.36(1.28) & 74.54(1.30) \\ 

\textbf{Haberman} & 73.35(3.32) & 72.03(2.67) & 73.06(2.97) & 72.76(3.29) & 74.22(2.85) & \textbf{76.13(2.06)} & 76.13(2.06) \\ 

\textbf{Heart} & 85.13(2.94) & \textbf{86.44(3.38)} & 85.62(3.03) & 85.13(2.75) & 84.15(4.35) & 85.46(2.70) & 85.30(2.30) \\ 

\textbf{ILPD} & 68.42(2.20) & 69.79(3.15) & 68.04(2.74) & 68.65(3.13) & 69.79(3.29) & 69.64(2.47) & \textbf{70.17(2.33)} \\ 

\textbf{Ionosphere} & 89.31(2.26) & \textbf{89.94(1.97)} & 88.80(2.60) & 89.56(2.20) & 89.94(2.48) & 89.06(2.21) & 89.06(2.21) \\ 

\textbf{Laryngeal1} & 86.16(4.00) & \textbf{87.42(2.98)} & 85.95(3.59) & 86.58(3.24) & 85.11(4.33) & 87.21(5.35) & 87.00(5.00) \\ 

\textbf{Laryngeal3} & \textbf{74.67(1.66)} & 73.67(2.14) & 74.17(2.25) & 73.79(2.03) & 71.67(3.34) & 73.54(1.66) & 73.42(1.26) \\ 

\textbf{Lithuanian} & 95.12(2.10) & 94.97(2.00) & 95.49(2.21) & 95.04(2.34) & \textbf{95.78(2.13)} & 93.26(3.22) & 93.12(3.09) \\ 

\textbf{Liver} & 71.50(4.96) & \textbf{72.02(4.72)} & 71.24(4.94) & 71.11(5.70) & 68.79(4.76) & 69.69(4.68) & 69.56(4.84) \\ 

\textbf{Magic} & 85.69 (1.37 ) & \textbf{86.02 (2.20)} & 85.79 (1.21) & 85.80(2.54) & 85.25(3.21) & 85.650(2.27) & 84.35(3.27) \\ 

\textbf{Mammographic} & 80.35(2.85) & 80.72(2.56) & 81.31(3.42) & 79.92(3.44) & 81.15(1.58) & 84.30(2.27) & \textbf{84.41(2.54)} \\ 

\textbf{Monk2} & 94.15(2.18) & \textbf{94.45(1.88)} & 94.35(1.72) & 94.45(1.88) & 92.91(1.84) & 83.45(3.46) & 83.34(3.32) \\ 

\textbf{Phoneme} & 84.76(0.77) & 85.05(1.08) & 84.62(0.95) & 85.16(1.16) & \textbf{85.22(0.88)} & 81.82(0.69) & 81.77(0.72) \\ 

\textbf{Pima} & 77.35(2.43) & 77.53(2.24) & 77.06(2.86) & 77.00(2.79) & \textbf{78.34(3.26)} & 77.93(1.86) & 77.76(1.75) \\ 

\textbf{Satimage} & 96.59(0.68) & \textbf{96.65(0.83)} & 96.50(0.82) & 96.55(0.80) & 96.46(0.78) & 96.46(0.79) & 96.42(0.76) \\ 

\textbf{Sonar} & 80.13(3.96) & 81.63(3.90) & 81.84(4.59) & 81.42(4.30) & \textbf{82.91(4.59)} & 82.06(5.09) & 81.84(5.67) \\ 

\textbf{Thyroid} & 96.60(1.12) & 96.99(0.75) & 96.60(0.77) & 96.86(0.91) & 96.60(0.77) & \textbf{97.38(0.67)} & 97.38(0.67) \\ 

\textbf{Vehicle} & 82.76(1.10) & 82.87(1.64) & 82.61(1.48) & 82.82(1.23) & 81.82(1.94) & 83.55(2.10) & \textbf{83.55(2.01)} \\ 

\textbf{Vertebral} & 85.47(3.21) & 84.90(5.33) & 85.05(4.71) & 84.90(6.15) & \textbf{86.47(2.38)} & 84.90(2.95) & 85.62(2.35) \\ 

\textbf{WDVG1} & 84.70(0.39) & 84.72(0.49) & 84.75(0.52) & 84.75(0.45) & 83.30(0.82) & 84.77(0.65) & \textbf{84.84(0.60)} \\ 

\textbf{Weaning} & 81.29(3.43) & \textbf{81.73(3.14)} & 80.86(3.75) & 81.44(3.23) & 80.71(3.89) & 79.98(3.55) & 79.69(3.71) \\ 

\textbf{Wine} & 99.02(1.61) & \textbf{99.52(1.11)} & 99.27(1.61) & 99.27(1.17) & 99.52(1.11) & 98.53(1.48) & 98.53(1.48) \\ 

\hline
\hline
\textbf{Average rank} & 3.80(0.78) & \textbf{3.00(0.92)}  & 4.03(0.90)  & 4.16(0.82) & 3.96(1.26) & 4.33(0.93) & 4.70(1.17) \\

\textbf{Win-Tie-Loss} & 17-3-10 &  19-9-2 &  16-4-10 & 17-2-11 & 15-1-14 & n/a & n/a \\

\textbf{Wilcoxon Signed Test}  & \scalebox{1.5}{\textasciitilde} ($\rho = .3044$) & \scalebox{1.5}{+} ($\rho = .0316$) & \scalebox{1.5}{\textasciitilde} ($\rho = .3389$) & \scalebox{1.5}{\textasciitilde} ($\rho = .2623$) & \scalebox{1.5}{\textasciitilde} ($\rho = .8612$) & n/a & n/a \\


\hline 
\end{tabular} } 
\end{table} 

\begin{figure*}[!ht]
  
   \begin{center}  	 
       	  \includegraphics[clip=,  width=0.900\textwidth]{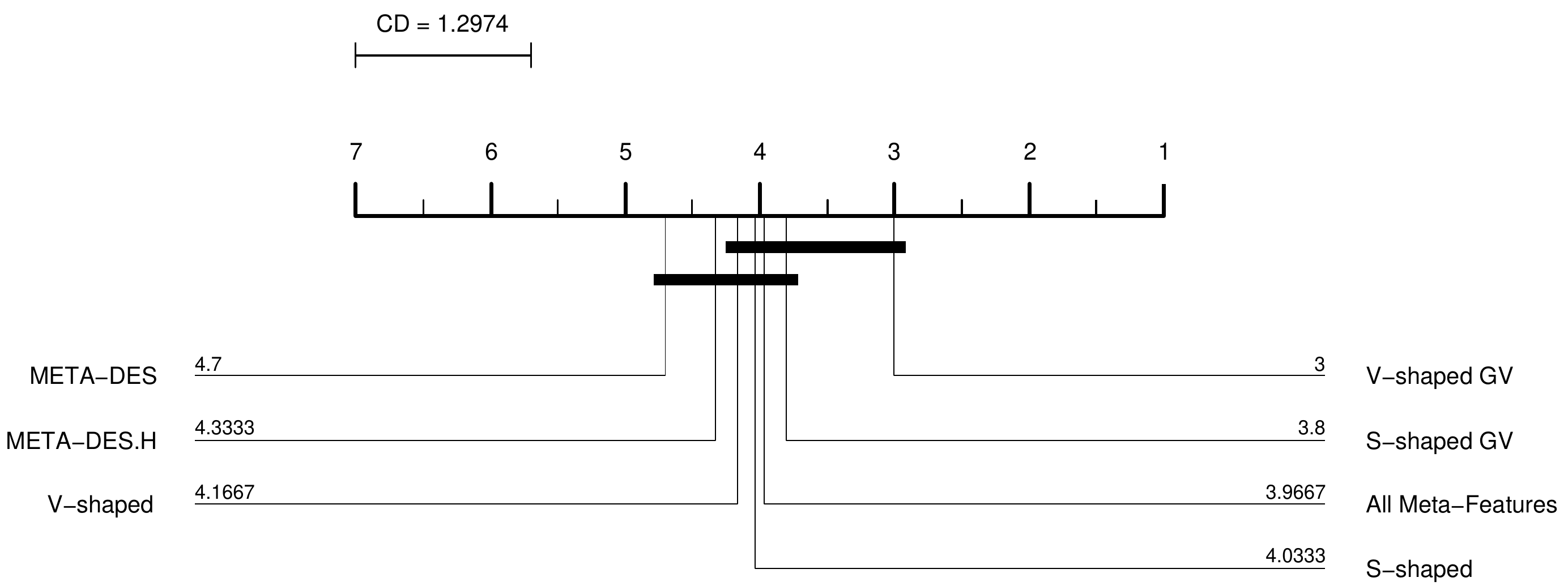}
   \end{center}
        \caption{Graphical representation of the average rank for each DES technique over the 30 datasets. For each technique, the numbers on the main line represent its average rank. The critical difference (CD) was computed using the Bonferroni-Dunn post-hoc test. Techniques with no statistical difference are connected by additional lines. }
        \label{fig:CDDiagramMETADESO}
\end{figure*}

Classification accuracies are reported in Table~\ref{table:resultsBPSO}. The best result achieved for each dataset is highlighted in bold. The Friedman~\cite{Friedman} test is used in order to compare the results of all techniques over the 30 classification datasets. The Friedman test is a non-parametric equivalent of the repeated ANOVA measures, used to make comparison between several techniques over multiple datasets~\cite{Demsar:2006}. For each dataset, the Friedman test ranks each algorithm, with the best performing one getting rank 1, the second best rank 2, and so forth. Then, the average rank and its standard deviation are computed, considering all datasets. The best algorithm is the one presenting the lowest average rank. Since we are comparing seven techniques, the degree of freedom is 6. We set the level of significance $\alpha = 0.05$, i.e., 95\% confidence. The Friedman test shows that there is a significant difference between the seven approaches. Then, a post-hoc Bonferroni-Dunn test was conducted for a pairwise comparison between the ranks achieved by each technique. The performance of two classifiers is significantly different if their difference in average rank is higher than the critical difference. The critical difference is computed using the following equation: $CD = q_{\alpha}\sqrt{\frac{k(k+1)}{6N}}$, where the critical value $q_{\alpha}$ is based on the Studentized range statistic divided by $\sqrt{2}$. The results of the post-hoc test are presented using the critical difference diagram proposed in~\cite{Demsar:2006} (Figure~\ref{fig:CDDiagramMETADESO}). The performance of techniques in which the difference in average ranks is higher than the critical difference are considered significantly different.  Techniques with no statistical difference are connected by a black bar in the CD diagram.

One interesting fact is that all techniques proposed in this work obtained lower rank values when compared to the previous version of the META-DES framework. The META-DES.Oracle using the V-shaped transfer function obtained the best overall performance, achieving an average rank of 3.00. Moreover, the results obtained by this technique were also significantly better than those obtained by both the META-DES and META-DES.H.

The second statistical analysis is conducted in a pairwise fashion in order to verify whether the difference in classification accuracy obtained by the META-DES.Oracle significantly improves the classification accuracy when compared to the previous versions of the framework. To that end, the Wilcoxon non-parametric signed rank test with the level of significance $\alpha = 0.05$ was used since it was suggested in~\cite{Demsar:2006} as a robust method for a pairwise comparison between classification algorithms over several datasets. The results of the Wilcoxon statistical test are shown in the last row of Table~\ref{table:resultsBPSO}. Techniques that achieve performances equivalent to the META-DES.H are marked with "\textasciitilde"; those that achieve statistically superior performance are marked with a "+", and those with inferior performance are marked with a "-". $\rho$-values are also shown in the last row of Table~\ref{table:resultsBPSO}.  

The results of the Wilcoxon signed rank test also demonstrate that the META-DES.Oracle using the V-Shaped transfer function and the global validation overfitting control scheme obtained classification results that are significantly superior when compared to both the META-DES.H and the META-DES, with a 95\% confidence over the 30 datasets considered in this work. Thus, based on the analysis, we can answer the two research questions posed at the beginning of this section: The meta-features selection optimization process does indeed significantly improve the classification performance of the system, when compared to the previous versions of the framework. In addition, we can also see that the system using 15 sets of meta-features without meta-feature selection, META-DES.ALL, achieves similar results when compared to previous versions of the framework (e.g., META-DES and META-DES.H). This suggest that simply adding more meta-features does not always lead to a better classification accuracy. The meta-feature selection stage is important for better addressing the behavior of the Oracle.

For the sake of simplicity, we refer to META-DES.Oracle, the version of the framework using the V-Shaped transfer function and global validation, in the rest of this paper.

\subsection{Comparison with the state-of-the-art DES techniques}

In this section, we compare the accuracy obtained by the proposed META-DES.Oracle against ten state-of-the-art dynamic selection techniques~\cite{Alceu2014}. The goal of this analysis is to know if the performance of the proposed system is significantly superior when compared to state-of-the-art DES techniques. The dynamic selection techniques used in this analysis are: Local Classifier Accuracy (LCA)~\cite{lca}, Overall Local Accuracy (OLA)~\cite{lca}, Modified Local Accuracy (MLA)~\cite{Smits_2002}, K-Nearest Oracles-Eliminate (KNORA-E), K-Nearest Oracles-Union (KNORA-U)~\cite{knora}, K-Nearest Output Profiles (KNOP)~\cite{paulo2}, Multiple Classifier Behavior (MCB)~\cite{mcb}, Randomized Reference Classifier (DES-RRC)~\cite{Woloszynski} and DCS-Rank~\cite{classrank}. These techniques were selected because they presented the very best results in the dynamic selection literature according to a recent survey on this topic~\cite{Alceu2014}. 

The same pool of classifiers is used for all techniques in order to ensure a fair comparison. For all techniques, the size of the region of competence, $K$, was set at $7$ since it achieved the best result in previous experiments~\cite{ijcnn2011,CruzPR}. The results are shown in Table~\ref{table:ResultsComparison}. For each dataset, we performed a pairwise comparison between the results obtained by the proposed META-DES.Oracle against those obtained by each state-of-the-art DES technique. The comparison was conducted using the Kruskal-Wallis non-parametric statistical test, with a 95\% confidence interval. Results that are significantly better are marked with a $\bullet$. In addition, the average rank of each technique, as well as the result of the sign test, are presented at the end of Table~\ref{table:ResultsComparison}.

 \begin{table*}[ht]
     \centering
     \caption{Mean and standard deviation results of the accuracy obtained for the proposed META-DES.Oracle and 10 state-of-the-art dynamic selection techniques. The best results are in bold. Results that are significantly better are marked with $\bullet$.} 
      \label{table:ResultsComparison} 
      \resizebox{1.0\textwidth}{!}{
      \begin{tabular}{|l  | c ||  c  c  c  c  c  c  c  c c c|}
     \hline
        \textbf{Database} & \textbf{META-DES.Oracle} & \textbf{KNORA-E}~\cite{knora} & \textbf{KNORA-U}~\cite{knora} & \textbf{DES-FA}~\cite{ijcnn2011} & \textbf{LCA}~\cite{lca} & \textbf{OLA}~\cite{lca} & \textbf{MLA}~\cite{Smits_2002} & \textbf{MCB}~\cite{mcb} & \textbf{KNOP}~\cite{paulo2} & \textbf{DES-RRC}~\cite{Woloszynski} & \textbf{DCS-Rank}~\cite{classrank}\\
         \hline
         \textbf{Adult} & \textbf{87.74(2.04)} $\bullet$ &  80.34(1.57) &  79.76(2.26) & 80.34(1.57) & 83.58(2.32) &  82.08(2.42) & 80.34(1.32)  & 78.61(3.32) & 79.76(2.26) & 86.71(1.53) & 83.04(2.42) \\	 
         
         \textbf{Banana} & 94.54(1.16) &  93.08(1.67) & 92.28(2.87) & \textbf{95.21(3.18)} & 95.21(2.15)  & 95.21(2.15)  & 80.31(7.20) &  88.29(3.38) & 90.73(3.45) & 86.44(1.76) & 93.44(1.73) \\
         
         \textbf{Blood} & \textbf{79.38(1.76)} $\bullet$ &  77.65(3.62) &  77.12(3.36) & 73.40(1.16) & 75.00(2.87) & 75.00(2.36) & 76.06(2.68)  & 73.40(4.19) & 77.54(2.03) & 75.89(1.41) & 74.35(2.49) \\
         
         \textbf{Breast (WDBC)} & 96.71(0.86) & 97.59(1.10)  &  97.18(1.02) & \textbf{97.88(0.78)} & 97.88(1.58)  & 97.88(1.58)  & 95.77(2.38)  & 97.18(1.38) & 95.42(0.89) & 96.71(0.61) & 96.01(1.00) \\
       
       \textbf{CTG} & \textbf{86.37(1.10)} & 86.27(1.57) &  85.71(2.20) & 86.27(1.57) &  86.65(2.35) &  86.65(2.35)  & 86.27(1.78) & 85.71(2.21) & 86.02(3.04) & 84.90(1.02) & 84.98(0.84) \\		
        
        \textbf{Ecoli} &  \textbf{81.57(3.47)} $\bullet$ &  76.47(2.76)  & 75.29(3.41) & 75.29(3.41) &  75.29(3.41)  &  75.29(3.41) &  76.47(3.06) & 76.47(3.06) & 80.00(4.25) & 78.82(3.58) & 76.73(3.52) \\	
        
        \textbf{Faults} & \textbf{69.32(1.18)} & 67.35(2.01) & 67.96(1.98)  & 68.17(1.59)  &  66.00(1.69) &  66.52(1.65)  & 67.76(1.54)  & 68.17(1.59) & 68.57(1.85) & 67.58(0.95) & 66.55(1.64)  \\	
        
        \textbf{Glass} & 66.46(4.22) &  57.65(5.85) &  61.00(2.88) & 55.32(4.98) & 59.45(2.65) & 57.60(3.65) & 57.60(3.65) & \textbf{67.92(3.24)} & 62.45(3.65) & 64.99(4.23) & 56.81(6.15) \\
         
         \textbf{German} & \textbf{76.58(1.99)} $\bullet$ & 72.80(1.95) & 72.40(1.80)  & 74.00(3.30) & 73.33(2.85) & 71.20(2.52) & 71.20(2.52) & 73.60(3.30)  & 73.60(3.30) & 75.07(2.36) & 69.78(2.70) \\	
         
         \textbf{Haberman} & 74.22(2.85)  &  71.23(4.16)  & 73.68(2.27)  & 72.36(2.41)  &  70.16(3.56) &  69.73(4.17) &  73.68(3.61)  &  67.10(7.65) & 75.00(3.40) & 75.15(2.50) & 70.32(4.06) \\        
         
         \textbf{Heart} & 86.44(3.38) &  83.82(4.05) & 83.82(4.05) & 83.82(4.05) &  85.29(3.69) & 85.29(3.69) & \textbf{86.76(5.50)} & 83.82(4.05) & 83.82(4.05) & 83.66(3.64) & 79.74(4.34) \\	
        
        \textbf{ILPD} & \textbf{69.79(3.15)}  &  67.12(2.35) &  69.17(1.58) &  67.12(2.35) & 69.86(2.20) & 69.86(2.20) & 69.86(2.20) &  68.49(3.27) & 68.49(3.27) & 67.88(1.89) & 67.81(2.52) \\	     
         
         \textbf{Ionosphere} & \textbf{89.94(1.97)} &  89.77(3.07) &  87.50(1.67) & 88.63(2.12) & 88.00(1.98)  & 88.63(1.98) & 81.81(2.52)  & 87.50(2.15) & 85.71(5.52) & 87.88(2.48) & 88.51(2.87) \\
         
         \textbf{Laryngeal1} & \textbf{87.42(2.98)} $\bullet$&  77.35(4.45) & 77.35(4.45) &  77.35(4.45) &  77.35(4.45) & 77.35(4.45) &  75.47(5.55) & 77.35(4.45) & 77.35(4.45) & 82.18(3.79) & 79.45(3.46) \\	

        \textbf{Laryngeal3} & \textbf{73.67(2.14) } & 70.78(3.68) & 72.03(1.89) & 72.03(1.89) &  72.90(2.30) & 71.91(1.01) & 61.79(7.80) & 71.91(1.01) & 73.03(1.89) & 72.41(1.87) & 66.67(6.13) \\

         \textbf{Lithuanian} & 94.97(2.00) &  93.33(2.50) &  95.33(2.64) & 98.00(2.46) & 85.71(2.20) & \textbf{98.66(3.85)} & 88.33(3.89)  & 86.00(3.33) & 89.33(2.29) & 85.04(1.57) & 93.41(1.22) \\
 
         \textbf{Liver} & \textbf{72.02(4.72)} $\bullet$ &  56.65(3.28) &  56.97(3.76) & 61.62(3.81) & 58.13(4.01)  & 58.13(3.27) & 58.00(4.25)  & 58.00(4.25) & 65.23(2.29)  & 63.70(4.14) & 61.24(5.42) \\

     	\textbf{Magic} & 86.02(2.20) &  80.03(3.25) & 79.99(3.55) & 81.73(3.27) & 81.53(3.35) & 81.16(3.00) & 73.13(6.35)  & 75.91(5.35) & 80.03(3.25) & \textbf{86.20(1.52)} & 76.72(1.13) \\
 		\textbf{Mammographic} & 80.72(2.56) & 82.21(2.27) & 82.21(2.27) & 80.28(3.02) & 82.21(2.27 & 82.21(2.27) & 75.55(5.50) & 81.25(2.07) & 82.21(2.27) & \textbf{84.29(1.32)} $\bullet$& 79.75(3.48) \\	
 		\textbf{Monk2} & \textbf{94.45(1.88)} $\bullet$ & 80.55(3.32) & 77.77(4.25) & 75.92(4.25)  & 74.07(6.60) & 74.07(6.60) & 75.92(5.65) & 74.07(6.60) & 80.55(3.32) & 80.86(2.58) & 86.21(4.93) \\	
         \textbf{Phoneme} & \textbf{85.05(1.08)} $\bullet$ &  79.06(2.50) &  78.92(3.33) &  79.06(2.50) & 78.84(2.53) &  78.84(2.53) &  64.94(7.75) & 73.37(5.55) & 78.92(3.33) & 73.64(1.55) & 79.45(0.88) \\	
          \textbf{Pima} & \textbf{77.53(2.24)} &  73.79(1.86) &  76.60(2.18) & 73.95(1.61) & 73.95(2.98)  & 73.95(2.56) & 77.08(4.56) & 76.56(3.71) & 73.42(2.11) & 75.41(2.73) & 72.97(2.25) \\
         \textbf{Satimage} & \textbf{96.65(0.83)}  $\bullet$&  95.35(1.23) & 95.86(1.07) & 93.00(2.90) & 95.00(1.40) & 94.14(1.07) & 93.28(2.10) & 95.86(1.07) & 95.86(1.07) & 95.60(0.75) & 94.76(0.97) \\	
         \textbf{Sonar} & \textbf{81.63(3.90)} $\bullet$ &  74.95(2.79) &  76.69(1.94) & 78.52(3.86) & 76.51(2.06) & 74.52(1.54) & 76.91(3.20)  & 76.56(2.58) & 75.72(2.82) & 80.13(5.09) & 79.27(5.67) \\
        \textbf{Thyroid} & \textbf{96.99(0.75)} $\bullet$ & 95.95(1.25) &  95.95(1.25) &  95.37(2.02) & 95.95(1.25) & 95.95(1.25) &  94.79(2.30) &  95.95(1.25) & 95.95(1.25) & 96.85(0.96) & 96.40(1.15) \\	
         \textbf{Vehicle} & 82.87(1.64) &  83.01(1.54) &  82.54(1.70)  & 82.54(4.05) & 80.33(1.84)  & 81.50(3.24) & 74.05(6.65) & \textbf{84.90(2.01)} & 80.09(1.47) & 82.76(1.81) & 79.61(1.97) \\
        \textbf{Vertebral} & 84.90(5.33) &  85.89(2.27) & \textbf{87.17(2.24)} &  82.05(3.20) &  85.00(3.25) &  85.89(3.74) &  77.94(5.80) & 84.61(3.95) & 86.98(3.21) & 85.90(3.68) & 83.62(3.38) \\		
       \textbf{WDGV1} & \textbf{84.72(0.49)} $\bullet$ &  84.01(1.10) &  84.01(1.10) & 84.01(1.10) & 80.50(0.56)  & 80.50(0.56) & 79.95(0.85)  & 78.75(1.35) & 84.21(0.45) & 84.46(0.48) & 83.85(0.61) \\	
        \textbf{Weaning} & 81.73(3.14) & 78.94(1.25) & 81.57(3.65) & \textbf{82.89(3.52)} & 77.63(2.35) & 77.63(2.35) & 80.26(1.52) & 81.57(2.86) & 82.57(3.33) & 78.51(3.29) & 77.19(2.18) \\
         \textbf{Wine} & \textbf{99.52(1.11)} $\bullet$ &  97.77(1.53) &  97.77(1.62) &  95.55(1.77)  & 85.71(2.25)  & 88.88(3.02) & 88.88(3.02)  & 97.77(1.62) & 95.50(4.14) & 98.52(1.57) & 92.10(5.57) \\
     	\hline
     	\hline
     	\textbf{Avg. rank} & \textbf{2.73(1.34)}  & 5.56(1.29) & 5.33(1.07) & 5.90(1.54) & 6.20(1.44) & 6.80(1.44) & 7.30(1.52) & 7.33(1.45) & 5.93(1.51) & 4.40(1.61) & 8.10(1.53) \\	

     	\textbf{Win-tie-loss} & n/a  & 4-3-23 & 6-1-23 & 6-1-23 & 5-2-23 & 7-1-22 & 3-2-25 & 5-1-24 & 5-0-25 & 4-2-24 & 1-0-29 \\	
     	
     	\textbf{Wilcoxon} & n/a  &  \scalebox{2}{-} ($\rho = .0003$) & \scalebox{2}{-} ($\rho = .0014$) & \scalebox{2}{-} ($\rho = .0014$)  & \scalebox{2}{-} ($\rho = .0152$) & \scalebox{2}{-} ($\rho = .0161$) & \scalebox{2}{-} ($\rho = .0001$) & \scalebox{2}{-} ($\rho = .0003$) & \scalebox{2}{-} ($\rho = .0003$) & \scalebox{2}{-} ($\rho = .0003$) & \scalebox{2}{-} ($\rho = .0003$) \\	

     \hline
     \end{tabular}
     }
 \end{table*}
 
  \begin{figure*}[!ht]
    
     \begin{center}  	 
         	  \includegraphics[clip=,  width=0.900\textwidth]{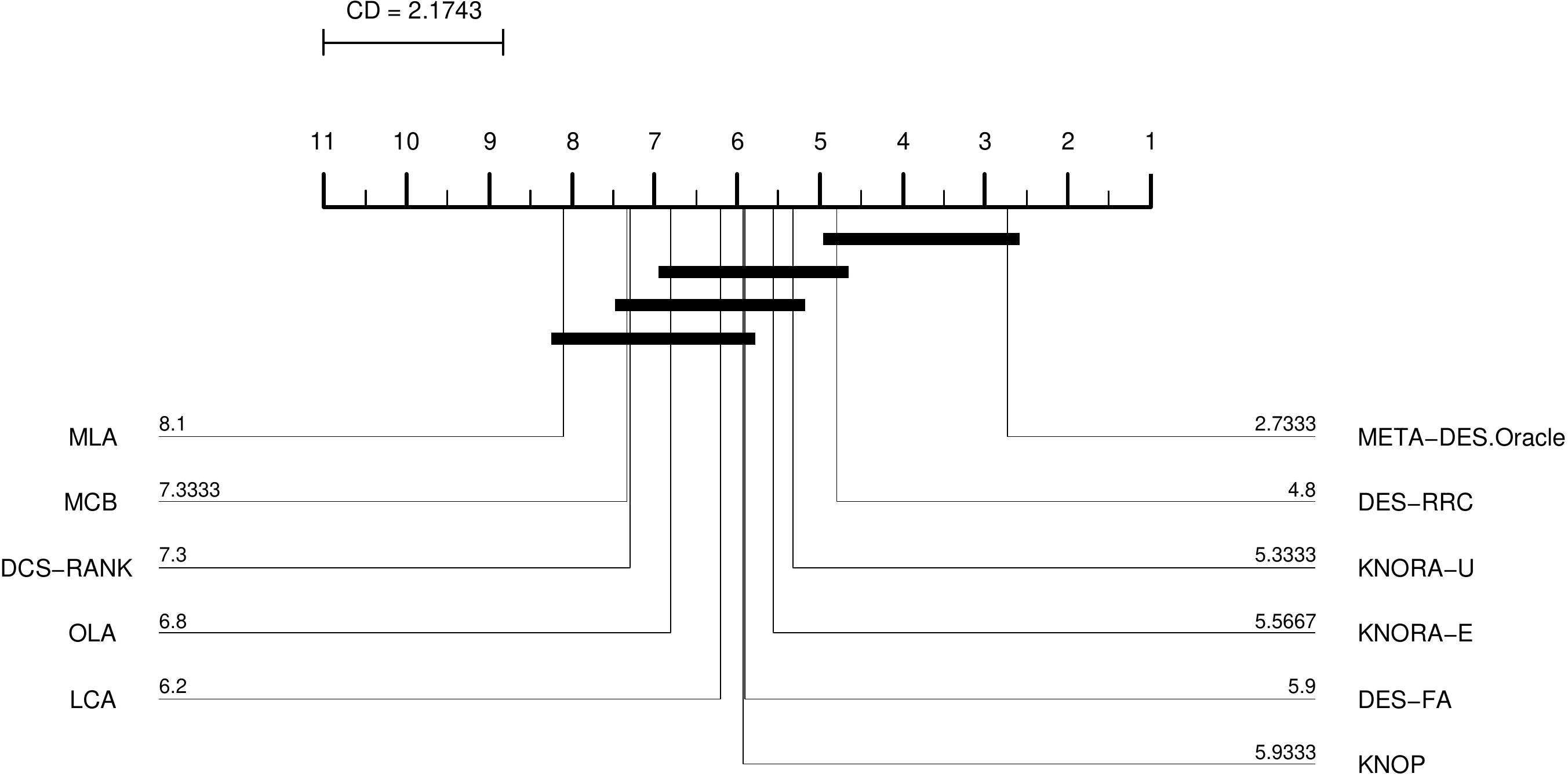}
     \end{center}
          \caption{Average rank of the dynamic selection methods over the 30 datasets. The best algorithm is the one presenting the lowest average rank.}
          \label{fig:CDDiagramDES}
  \end{figure*}

Figure~\ref{fig:CDDiagramDES} illustrates the average rank of each technique using the CD diagram. Similarly to Section~\ref{sec:comparative}, the CD was calculated using the Bonferroni-Dunn post-hoc test. The  META-DES.Oracle obtained the lowest average rank, 2.73, followed by the technique based on probabilistic models, DES-RRC~\cite{Woloszynski}, presenting an average rank of 4.40. Hence, the performance of the META-DES.Oracle is significantly better when compared to the majority of the state-of-the-art DES techniques. Only the DES-RRC obtained a statistically equivalent performance. However, when we compared those two techniques in terms of wins, ties and losses as reported in Table~\ref{table:ResultsComparison}, we could see that the META-DES.Oracle obtained the best accuracy for 24 datasets, while the DES-RRC outperformed the META-DES.Oracle only in 4 datasets. For two datasets, the results of both techniques were tied. Furthermore, we also performed the Wilcoxon non-parametric signed rank test with the level of significance $\alpha = 0.05$ for a pairwise comparison between the results obtained by the META-DES.Oracle against state-of-the-art DES techniques over the 30 datasets. The results of the Wilcoxon test are presented in the last row of Table~\ref{table:ResultsComparison}.  
 
When a pairwise comparison between the techniques is performed, we can see that the META-DES.Oracle dominates when compared against previous DES techniques. Its performance is statistically better when compared to any of the 10 state-of-the-art techniques. This can be explained by two factors: state-of-the-art DES techniques are based only on one criterion to estimate the competence of the base classifier; this could be, local accuracy, ranking, probabilistic models, etc. For instance, the ranking and probabilistic criteria used by the DCS-Rank and DES-RRC techniques are embedded in the META-DES framework as meta-features $f_{rank}$ and $f_{PRC}$, respectively. In addition, through the BPSO meta-features selection scheme, only the meta-features that are relevant for the given classification problem are selected and used for the training of the meta-classifier. As shown in Figure~\ref{fig:selectedFeatures}, the selected meta-features vary considerably according to different classification problems. Thus, it is expected that the proposed framework obtains a significant gain in performance when compared to previous DES techniques.
  
\subsection{Comparison with Static techniques}
In this section, we compare the results obtained by the META-DES.Oracle against static ensemble techniques as well as single classifier models. For the static ensemble techniques, we evaluate the performance of the AdaBoost~\cite{boosting}, Bagging~\cite{bagging}, Random Forest~\cite{breiman2001random,rokach2016decision}, the classifier with the highest accuracy in the validation data (Single Best) and a static ensemble selection method based on the majority voting error proposed in~\cite{classmaj}. Furthermore, two single classifier models were considered: Multi-Layer Perceptron (MLP) Neural Network and a Support Vector Machine with Gaussian Kernel (SVM). These classifiers were selected based on a recent study~\cite{delgado14a} that ranked the best classifiers in a comparison considering a total of 179 classifiers over 121 classification datasets. 

The objective of this study is to determine whether the proposed META-DES.Oracle obtain recognition accuracy that is either statistically better or equivalent to the ones achieved by the best classifiers in the literature~\cite{delgado14a}. This is an important analysis since the DES literature still lacks a comparison with classical classification approaches that do not use ensembles. In the DES literature, the accuracy of the proposed techniques are only compared either with other DES techniques or with static ensemble selection considering the same pool of classifiers~\cite{Alceu2014}. 

All classifiers were evaluated using the Matlab PRTOOLS toolbox~\cite{PRTools}. Since static techniques require neither a meta-training nor a dynamic selection phase, the training ($\mathcal{T}$) and meta-training set ($\mathcal{T}_{\lambda}$) were merged into a single training set. The dynamic selection dataset (DSEL) was used as the validation dataset. The test set, $\mathcal{G}$, remained unchanged. For each replication, the hyper-parameters of the each classifier model were set as follows:
 
\begin{enumerate} 
	
\item MLP Neural Network: We varied the number of neurons in the hidden layer from 10 to 100 at 10 point intervals. The configuration that achieved the best results in the validation data was used. The MLP training process was conducted using the Levenberg-Marquadt algorithm. The process was stopped if the performance on the validation set decreased or failed to improve for five consecutive epochs.

\item SVM with a Gaussian Kernel: A grid search was performed in order to set the values of the regularization parameter, $c$, and the Kernel spread parameter $\gamma$. 

\item Random Forest: We varied the number of trees from 25 to 200 at 25 point intervals. The configuration with the highest performance on the validation dataset was used for generalization.  

\end{enumerate}

\begin{figure*}[!ht]

   \begin{center}  	 
       	  \includegraphics[clip=,  width=0.900\textwidth]{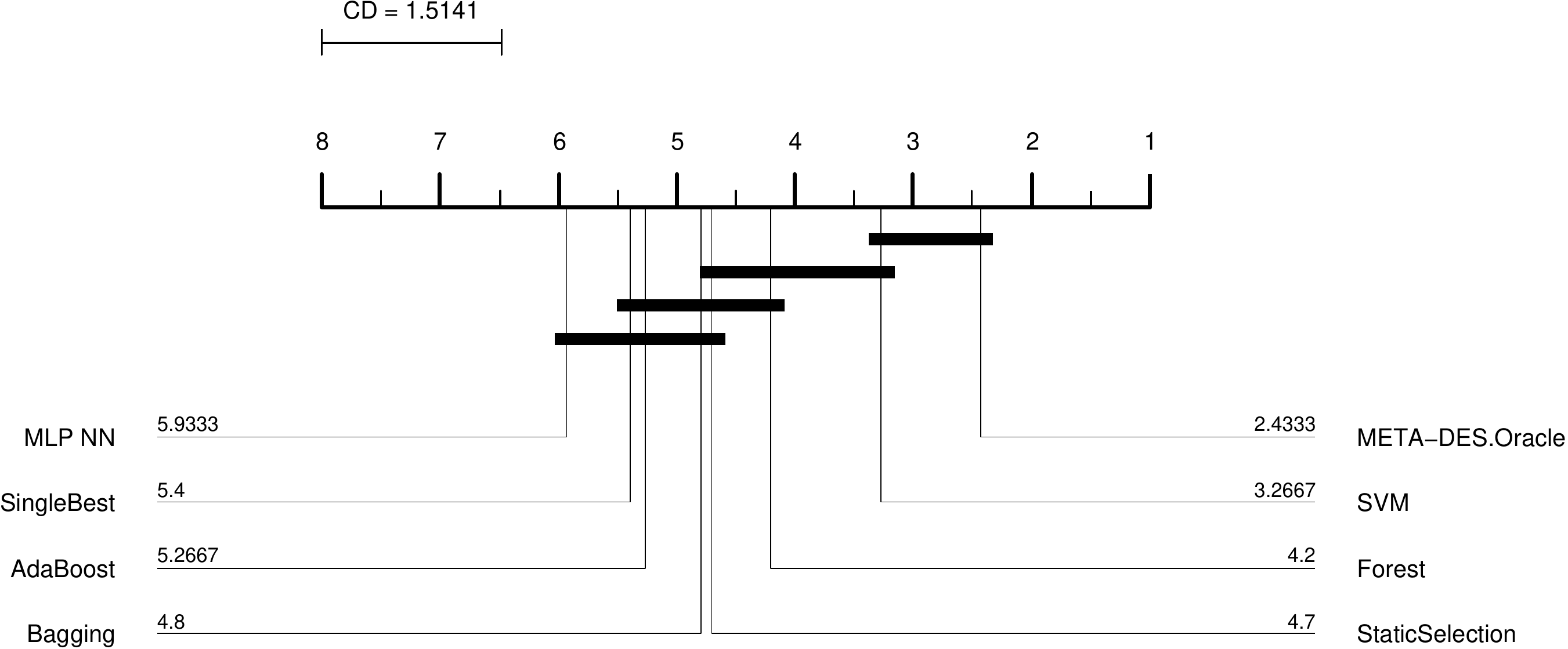}
   \end{center}
        \caption{Average rank of the dynamic selection methods over the 30 datasets. The best algorithm is the one presenting the lowest average rank.}
        \label{fig:CDDiagramStatic}
\end{figure*}

The classification accuracy of each technique is reported in Table~\ref{table:Results2}. For each dataset, we performed a pairwise comparison between the results obtained by the proposed META-DES.Oracle, against the results obtained by each state-of-the-art DES technique. The comparison was conducted using the Kruskal-Wallis non-parametric statistical test, with a 95\% confidence interval. Results that are significantly better at a 95\% confidence are marked with $\bullet$. Moreover, we also report the average ranks and the results of the Wilcoxon test at the end of Table~\ref{table:Results2}. Figure~\ref{fig:CDDiagramStatic} illustrates the critical difference diagram.

\begin{table}[!ht]
    \centering
    \caption{Mean and standard deviation results of the accuracy obtained for the proposed META-DES and static classification models. The best results are in bold. Results that are significantly better ($p < 0.05$) are marked with $\bullet$.}
     \label{table:Results2} 
     \resizebox{1.0\textwidth}{!}{
     \begin{tabular}{|c | c || c c c c c c c|}
    \hline

      \textbf{Database} & \textbf{META-DES.Oracle} & \textbf{Single Best~\cite{Alceu2014}} &  \textbf{Bagging}~\cite{bagging} & \textbf{AdaBoost}~\cite{boosting} & \textbf{Static Selection~\cite{classmaj}} & \textbf{MLP NN} & \textbf{SVM} & \textbf{Random Forest} \\
        \hline
        \textbf{Adult} & \textbf{87.74(2.04)} $\bullet$&  83.64(3.34) &  85.60(2.27) & 83.58(2.91) & 84.37(2.79) & 80.33(3.25) & 85.31(3.06) & 83.03(4.60)  \\	 
        
        \textbf{Banana} & 94.54(1.16) & 84.07(2.22) & 81.43(3.92) & 81.61(2.42) & 81.35(4.28) & 98.11(0.85) & \textbf{98.19(0.78)} $\bullet$ & 97.02(1.03)  \\
        
        \textbf{Blood} & \textbf{79.38(1.76)} $\bullet$ & 75.07(1.83) & 75.24(1.67) & 75.18(2.08) & 75.74(2.23) & 76.38(1.48) & 75.42(4.23) & 73.03(6.35)  \\
        
        \textbf{Breast (WDBC)} & 96.71(0.86)  & 97.04(0.74) & 96.35(1.14) & \textbf{98.24(0.89)} $\bullet$ & 96.83(1.00) & 95.77(0.74) & 97.81(1.07) & 95.85(1.37)  \\
	    
	    \textbf{CTG} & 86.37(1.10) &  84.21(1.10) & 84.54(1.46) & 83.06(1.23) & 84.04(2.02) & 88.19(2.27)  & \textbf{92.29(0.76)} $\bullet$ & 91.27(1.20)  \\		
       
       \textbf{Ecoli} & 81.57(3.47) &  69.35(2.68) & 72.22(3.65) & 70.32(3.65) & 67.80(4.60) & 74.35(14.08) & \textbf{83.88(2.42)} & 67.65(7.55)  \\	
       
       \textbf{Faults} & 69.32(1.18) &  66.05(1.98) &  67.02(1.98) & 66.57(1.06) & 67.22(1.64) & 68.99(2.63) &  \textbf{74.00(1.72)} $\bullet$ & 69.83(3.05)  \\     
       
       \textbf{Glass} & 66.46(4.22) &  52.92(4.53) &  62.64(5.61) & 55.89(3.25) & 57.16(4.17) & 56.22(7.99) & 60.60(5.17) & \textbf{66.54(6.01)}  \\     
        
        \textbf{German credit} & \textbf{76.58(1.99)} $\bullet$ &  71.16(2.39) &  74.76(2.73) & 72.96(1.25) & 73.60(2.69) & 64.20(3.98) & 75.32(1.70) & 70.35(5.85) \\
        
        \textbf{Haberman} & 74.52(2.94)  & \textbf{75.65(2.68)} & 72.63(3.45) & 75.26(3.38) & 73.15(3.68) & 68.42(5.15) & 71.10(2.21) & 63.81(7.23)  \\        
        
        \textbf{Heart} & \textbf{86.44(3.38)} $\bullet$ &  80.26(3.58) &  82.50(4.60) & 81.61(5.01) & 82.05(3.72) & 71.17(6.86) & 83.44(3.28) & 77.79(3.27)  \\	
        
        \textbf{ILPD} & \textbf{69.79(3.15)} &  67.53(2.83) &  67.20(2.35) & 69.38(4.28) & 67.26(1.04) & 64.31(3.68) & 66.23(3.95) & 65.68(3.94)  \\	        
        
        \textbf{Ionosphere} & 89.94(1.97) & 87.29(2.28) & 86.75(2.75) & 86.75(2.34) & 87.50(2.23) & 86.36(4.31) & \textbf{94.54(1.58)} $\bullet$ & 94.09(2.50)  \\
        
        \textbf{Laryngeal1} & \textbf{87.42(2.98)} $\bullet$&  80.18(5.51) &  81.32(3.82) & 79.81(3.88) & 80.75(4.93) & 76.98(6.01) & 
        81.69(4.70) & 80.18(4.81)  \\	

	    \textbf{Laryngeal3} & 73.67(2.14) &  68.42(3.24) &  67.13(2.47) & 62.32(2.57) & 71.23(3.18) & 64.26(4.19) & \textbf{74.60(2.95)} & 71.12(4.73)  \\	        
        
        \textbf{Lithuanian} & 94.97(2.00) & 84.35(2.04) & 82.33(4.81) & 82.70(4.55) & 82.66(2.45) & 92.66(3.15) & \textbf{96.40(1.70)} $\bullet$& 95.53(1.50)  \\        
        
        \textbf{Liver} & \textbf{72.02(4.72)} & 65.38(3.47)  & 62.76(4.81) & 64.65(3.26) & 59.18(7.02) & 61.86(4.86) & 71.27(4.10) & 67.32(4.79) \\                             
    	
    	\textbf{Magic} & 86.02(2.20) &  80.27(3.50) &  81.24(2.22) & \textbf{87.35(1.45)} & 85.25(3.25) & 83.07(2.20) & 87.20(1.52) & \textbf{88.65(2.32)}  \\	  
		
		\textbf{Mammographic} & 80.72(2.56) &  83.60(1.85) & \textbf{85.27(1.85)} $\bullet$ & 83.07(3.03) & 84.23(2.14) & 77.88(9.87) & 80.29(1.83) & 77(1.12)  \\	
		
		\textbf{Monk2} & 94.45(1.88)  &  79.25(3.78) & 79.18(2.57) & 80.27(2.76) & 80.55(3.59) & \textbf{99.25(1.21)} $\bullet$ & 96.57(1.38) & 83.88(3.09)  \\	
        
        \textbf{Phoneme} & 85.05(1.08) & 75.87(1.33) &  72.60(2.33) & 75.90(1.06) & 72.70(2.32) & 82.11(4.17) & 76.27(1.85) & \textbf{89.59(0.20)} $\bullet$ \\  
        
        \textbf{Pima} & \textbf{77.53(2.24)} & 73.57(1.49) & 73.28(2.08) & 72.52(2.48) & 72.86(4.78) & 69.37(2.94) & 76.56(2.71) & 74.32(3.92)  \\
        
        \textbf{Satimage} & \textbf{96.65(0.83)} &  94.52(0.96) & 95.23(0.87) & 95.43(0.92) & 95.31(0.92) & 92.65(2.97) & 91.15(1.20) & 96.21(1.42)  \\	
        
        \textbf{Sonar} & 81.63(3.90) & 78.21(2.36) & 76.66(2.36) & 74.95(5.21) & 79.03(6.50) & 76.15(6.09) & 82.80(3.99) & \textbf{84.80(6.62)} $\bullet$  \\
       
       \textbf{Thyroid} & \textbf{96.99(0.75)} $\bullet$&  95.15(1.74) & 95.25(1.11) & 96.01(0.74) & 96.24(1.25) & 94.98(1.35) & 94.79(0.10) & 95.08(0.49)  \\	
        
        \textbf{Vehicle} & \textbf{82.87(1.64)} $\bullet$& 81.87(1.47) & 82.18(1.31) & 80.56(4.51) & 81.65(1.48) & 72.31(8.63) & 74.19(3.00) & 79.00(2.42)  \\
       
       \textbf{Vertebral Column} & 84.90(5.33) &  82.04(2.17) &  85.89(3.47) & 83.22(3.59) & 84.27(3.24) & 84.14(4.55) & 84.74(4.33) & 84.48(3.93)  \\	
       
       \textbf{WDG V1} & 84.72(0.49) &  83.17(0.76) &  84.36(0.56) & 84.04(0.37) & 84.23(0.53) & 81.68(7.82) & \textbf{86.90(0.09)} $\bullet$ & 85.89(0.46) \\	
       
       \textbf{Weaning} & 81.73(3.14) &  74.86(4.78) &  76.31(4.06) & 74.47(3.68) & 76.89(3.15) & 80.92(4.77) & 87.23(1.96) & \textbf{88.25(2.93)} $\bullet$ \\
        
        \textbf{Wine} & \textbf{99.52(1.11)} $\bullet$ & 96.70(1.46) & 95.56(1.96) & 99.20(0.76) & 96.88(1.80) & 92.88(10.30) & 98.88(1.17) & 97.33(2.29)  \\

    	\hline
    	\hline
    	\textbf{Average rank} & \textbf{2.43(0.86)} &  5.40(0.87) &  4.80(1.02) & 5.26(1.03) & 4.70(0.83) & 4.93(1.16) & 3.26(1.04) & 4.20(1.28)  \\

    	\textbf{Win-Tie-Loss} & n/a &  3-1-26 &  4-0-26 & 5-1-24 & 3-1-26 & 4-1-25 & 12-3-15 & 10-2-18  \\
    	
    	\textbf{Wilcoxon Signed Test} & n/a  & \scalebox{2}{-} ($\rho = .0001 $) & \scalebox{2}{-} ($\rho = .0001 $)  & \scalebox{2}{-} ($\rho = .0003$)  & \scalebox{2}{-} ($\rho = .0001$) & \scalebox{2}{-} ($\rho = .0101$)  & \scalebox{2}{\textasciitilde} ($\rho = .3600$) & \scalebox{2}{\textasciitilde} ($\rho = .2005$)  \\	
    \hline
    \end{tabular}
} 
\end{table}

Based on the result,s we can conclude that the META-DES.Oracle outperforms static ensemble techniques. This result was expected since many works in the DES literature have shown that dynamic selection outperforms static combination rules in many applications~\cite{Alceu2014}. Moreover, this claim is especially true when a pool of weak linear classifiers is considered since they become experts into different regions of the feature space. As reported in~\cite{reportarXiv}, a static combination of base classifiers in such a case may not yield a good classification performance since there may never be a consensus in the correct answer between the classifiers in the pool. However, when dynamic selection is used, only the most competent classifiers for the given query sample are selected to predict its label. As such, the classifiers that are not experts in the local region do not influence the ensemble decision negatively.

When compared with single classifier models, the META-DES.Oracle obtained the lowest average rank. The results achieved META-DES.Oracle is statistically equivalent to those achieved by the SVM classifier, based on both the Friedman test with Bonferroni-Dunn post-hoc test, and the Wilcoxon sign test at $\alpha = 0.05$ significance. Hence, the analysis demonstrate the classification performance achieved by the proposed META-DES.Oracle is among the best classifier models in the literature, since both SVM and Random Forests presented the overall best performance in the analysis conducted by Delgado et al.~\cite{delgado14a}. 

It is important to point out that the META-DES.Oracle obtained a small advantage in terms of wins, ties and losses when compared to the SVM classifier. The META-DES.Oracle presented the best recognition accuracy in 16 datasets, while the SVM obtained a higher accuracy in 12 datasets. For two datasets (Vertebral Column and Mammographic), the results were tied. This result may be explained by the fact most of the datasets used in this analysis are ill-defined, i.e., small sample size datasets. For such datasets, the training data may not have enough samples to train a single classifier model and select the best hyperparameters, e.g., the number of neurons in the hidden layer of an MLP neural network, or the regularization parameter, $c$, and the Kernel spread parameter, $\gamma$, in an SVM. In addition, since the training set was small, there might be variations between the training and test distribution. The META-DES.Oracle obtained the best results for several ill-defined problems, such as Liver disorders, Blood transfusion, Heart, Laryngeal1, Wine and Thyroid. Those are all small-sized datasets with less than 500 samples available for training. One advantage of the META-DES framework is that the pool is composed of linear classifiers which do not require the selection of any hyper-parameters. Thus, the training can be performed using small size datasets. Since the training set is relatively small, the classifiers may specialize in local regions of the feature space. Using dynamic selection, only the most competent classifiers in the local region where the test sample is located are used to predict its label. Thus, through DES, it is still possible to obtain high classification accuracy even for ill-defined problems. 

Furthermore, the optimization process of the META-DES.Oracle framework is conducted in the meta-problem, using the meta-data extracted in the meta-training stage. Several meta-feature vectors are generated for each training sample in the meta-training phase. For instance, consider that $200$ training samples are available for the meta-training stage ($N = 200$); if the pool $C$ is composed of $100$ weak classifiers ($M = 100$), the meta-training dataset is the number of training samples $N$ $\times$ the number classifiers in the pool $M$, $N \times M = 20.000$. Hence, even though the problem may be ill-defined, the framework generates enough meta-training data in order to properly train the meta-classifier. There is more data to train the meta-classifier $\lambda$ than for the generation of the pool of classifiers $C$ itself. Hence, even though the classification problem may be ill-defined, given the size of the training set, using the proposed framework, we can overcome this limitation since the size of the meta-problem is up to 100 times bigger than the classification problem.

\section{Conclusion}
\label{sec:conclusion}

In this paper, we propose a novel DES framework using meta-learning and Oracle information, called META-DES.Oracle. 15 sets of meta-features are proposed, using different sources of information found in the DES literature for dynamically estimating the level of competence of base classifiers; these include, local accuracy, ranking, probabilistic, ambiguity and behavior. Next, a meta-feature selection scheme using overfitting cautious Binary Particle Swarm Optimization is performed to optimize the performance of the meta-classifier. The optimization process is guided by a formal definition of the Oracle. Thus, the meta-classifier can better address the complex behavior of the Oracle.

We have conducted a case study using the P2 problem, which is a synthetic dataset with a complex non-linear decision border. We demonstrate that using a pool composed of 5 linear Perceptron classifiers, it is possible to approximate the complex decision boundary of the P2 problem using the proposed framework. The proposed META-DES.Oracle obtained a recognition performance of 97\%, which is closer to the results obtained by the Oracle, and compares very favorably against previous versions of the META-DES framework.

Experiments were conducted using 30 classification problems. First, we performed an analysis of the meta-features that were selected for each problem. The analysis demonstrated that the selected sets of meta-features varies considerably according to different datasets. In addition, each meta-feature was selected in at least 20\% of the datasets. All sets of meta-features was thus relevant in better addressing the complex behavior of the Oracle. Next, the performance obtained by the proposed META-DES.Oracle was compared with previous versions of the META-DES framework, as well as ten state-of-the-art dynamic selection techniques. Experimental results demonstrate that the META-DES.Oracle outperforms the previous versions of the technique in the majority of the datasets. In addition, the gain in performance obtained by the META-DES.Oracle is shown to be statistically significant based on both the Friedman test with a post-hoc Bonferroni-Dunn correction and the Wilcoxon sign rank test. Thus, the BPSO meta-features selection scheme proposed in this paper does indeed significantly improve the classification performance of the framework.

When compared with static and single classifier methods, the results achieved by the proposed META-DES.Oracle are comparable with the best performing classifiers.  Moreover, the results confirm the claim that DES techniques outperform single classifier models for ill-defined problems. Since the optimization process of the META-DES.Oracle is performed using the meta-data generated during the meta-training stage, there is enough data to train and optimize the meta-classifier. Thus, the proposed framework can deal with small sample size classification problems.

\section*{Acknowledgment}
This work was supported by the Natural Sciences and Engineering Research Council of Canada (NSERC), the \'{E}cole de technologie sup\'{e}rieure (\'{E}TS Montr\'{e}al) and CNPq (Conselho Nacional de Desenvolvimento Cient\'{i}fico e Tecnol\'{o}gico).

\bibliographystyle{elsarticle-num}
\bibliography{report}

\end{document}